\documentclass[10pt]{article} % For LaTeX2e
\usepackage[preprint]{tmlr}
\usepackage{amsmath,amsfonts,bm}

\def\eqref#1{equation~\ref{#1}}
\def\1{\bm{1}}

\DeclareMathAlphabet{\mathsfit}{\encodingdefault}{\sfdefault}{m}{sl}
\SetMathAlphabet{\mathsfit}{bold}{\encodingdefault}{\sfdefault}{bx}{n}

\usepackage{hyperref}
\usepackage{url}
\usepackage[T1]{fontenc}
\usepackage{amsmath,amssymb,amsthm,graphicx,booktabs}
\usepackage{xcolor,url,etoolbox}
\usepackage{float,needspace,tabularx,array,placeins}
\usepackage[most]{tcolorbox}
\usepackage{enumitem}
\usetikzlibrary{arrows.meta,positioning,calc}
\newenvironment{discoverychart}{\begin{tikzpicture}[
  finding/.style={draw=linkblue!65!white,fill=white,rounded corners=2pt,line width=.5pt,text width=.44\linewidth,minimum height=12mm,align=center,inner xsep=5pt,inner ysep=5pt,font=\small},
  transition/.style={-{Stealth[length=2mm]},draw=linkblue,line width=.7pt},
  operation/.style={right=2mm,text width=.43\linewidth,align=left,font=\footnotesize,text=black}]}{\end{tikzpicture}}
\newcommand{\discoveryfirst}[2]{\node[finding] (#1) {#2};}
\newcommand{\discoverynext}[4]{\node[finding,below=6mm of #2] (#1) {#3};\draw[transition] (#2.south) -- (#1.north);\coordinate (step-#1) at ($(#2.south)!.5!(#1.north)$);\draw[draw=linkblue!45!white,line width=.4pt] (step-#1) -- ++(.26\linewidth,0) node[operation] {#4};}
\usepackage{hyperref}
\newcolumntype{Y}{>{\raggedright\arraybackslash}X}
\definecolor{linkblue}{HTML}{245B77}
\newtcolorbox{pipelineoutput}{enhanced,breakable,
  colback=linkblue!3!white,colframe=linkblue!65!white,
  colbacktitle=linkblue!10!white,coltitle=linkblue!85!black,
  title={Automatically generated by the PatternMiner system},
  title after break={PatternMiner system output (continued)},
  fonttitle=\small\bfseries,boxrule=0.5pt,arc=1pt,boxsep=0pt,
  left=8pt,right=8pt,top=7pt,bottom=7pt,toptitle=5pt,bottomtitle=5pt,
  grow to left by=8.5pt,grow to right by=8.5pt,
  before skip=10pt,after skip=10pt}
\hypersetup{colorlinks=true,linkcolor=linkblue,citecolor=linkblue,urlcolor=linkblue,
pdftitle={PatternMiner: Automated Mathematical Discovery Through LLM-Guided Training and Interpretation of Small Models}}
\newcommand{\Des}{\operatorname{Des}}
\newcommand{\Fl}{\operatorname{Fl}}
\newcommand{\C}{\mathbb{C}}
\newcommand{\Z}{\mathbb{Z}}

\newtheorem{theorem}{Theorem}[subsection]
\newtheorem{lemma}[theorem]{Lemma}
\newtheorem{corollary}[theorem]{Corollary}
\newtheorem{proposition}[theorem]{Proposition}
\newtheorem{conjecture}[theorem]{Conjecture}

\title{\raggedright PatternMiner: Automated Mathematical Discovery Through LLM-Guided Training and Interpretation of Small Models}

\author{\name Michael Ruofan Zeng, Junaid Hasan, \\
      \name Xuanyu Yang, Emily Xia Meng \hfill \email zengrf@uw.edu \\
      \addr University of Washington
      \AND
      \name Henry Kvinge \email henry.kvinge@pnnl.gov \\
      \addr Pacific Northwest National Laboratory \\
      University of Washington}

\def\month{MM}  % Insert correct month for camera-ready version
\def\year{YYYY} % Insert correct year for camera-ready version
\def\openreview{\url{https://openreview.net/forum?id=XXXX}} % Insert correct link to OpenReview for camera-ready version

\begin{document}

\maketitle

\begin{abstract}
There are several approaches to AI-driven mathematical discovery. One of these involves the use of large language models (LLMs) to resolve well-specified conjectures via text-based reasoning, while another uses narrow models and interpretability for more open-ended, human-in-the-loop exploration using object-based mathematical data. The latter is limited in its ability to move beyond simple empirical pattern detection while the former is not well-designed for direct, open-ended exploration of large mathematical datasets. In this paper we introduce \emph{PatternMiner}, a system that combines these two paradigms. In PatternMiner, an LLM agent generates mathematical datasets based on an initial broad question, trains narrow machine learning models on them, uses interpretability techniques to extract patterns from these trained models, and uses these patterns to formulate conjectures. Once an initial conjecture has been formulated, PatternMiner instructs the agent to search for counterexamples using calls to mathematical software. If a set of counterexamples can be found, these are used as feedback for conjecture refinement or another round of training and interpretability. The conjecture progressively evolves through this iterative sequence with the agent allowed to try to prove it at any point. We describe the results of applying PatternMiner to five well-known positivity problems in algebraic combinatorics. PatternMiner proves several new theorems related to Schubert positivity and projection, a character-row evaluation formula, a restricted version of the Kazhdan--Lusztig combinatorial invariance conjecture, and Schur positivity for a family of immanant coefficients. We end by analyzing the most common routes to conjecture generation that are used by PatternMiner in our case studies as well as documenting the most common failure modes.
\end{abstract}

\section{Introduction}

Large language models (LLMs) are now capable of solving well-defined, research-level mathematical conjectures. Examples include the three-dimensional counterexample to the Jacobian conjecture found by Alp\"oge with Claude \citep{ramos2026jacobian} and an announced proof of finite-time blow-up for the forced Navier--Stokes equations \citep{openai2026navierstokes,bastounis2026translation}. However, formulating conjectures is often just as important as proving them. For human mathematicians, conjectures arise through interplay between theory-based reasoning and detection of patterns within examples. Current AI-based approaches to conjecture generation emphasize just one of these, with LLM-based approaches focusing on theory-based reasoning \citep{onda2025leanconjecturer,meca2026}. Another family of approaches uses mechanistic interpretation on small models trained on a specific task to drive conjecture generation \citep{davies2021,chau2025}, although they can be labor-intensive. As mathematics moves from ``proof scarcity'' toward ``proof abundance'' \citep[Section~7]{tao2026mathai}, ``conjecture scarcity'' makes it increasingly important to discover good conjectures at scale.

\begin{figure}[!htbp]
\centering
\includegraphics[trim=0 4.5bp 0 4.5bp,clip,width=.88\linewidth]{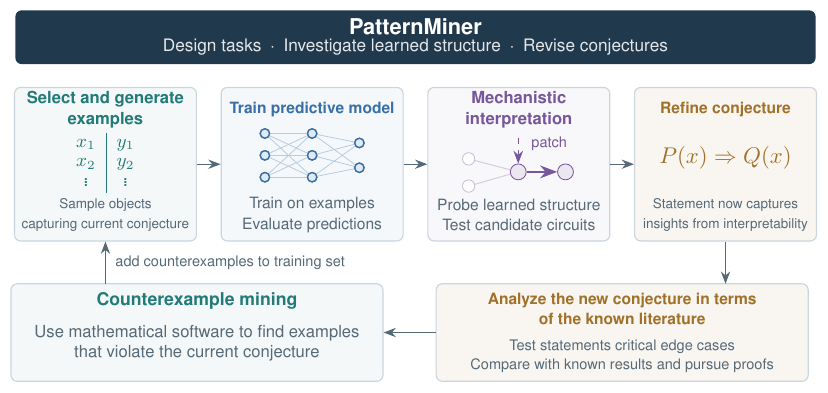}
\caption{A diagram of the PatternMiner framework.}
\label{fig:pipeline}
\end{figure}

We introduce PatternMiner\footnote{Code and experiment records are available in the repository: \url{https://github.com/uw-math-ai/PatternMiner}}, a system for mathematical discovery that combines these two approaches. In PatternMiner an LLM agent designs and runs the experiments, drawing on its broad mathematical knowledge to ground insights derived from the interpretation of small, specialist models trained on narrow, well-defined tasks. To begin, the agent designs a training task in line with the area of inquiry and then generates data related to this task using mathematical software like SageMath \citep{sagemath}. The agent then trains and interprets a small model such as a linear scorer, decision stump, boosted tree model, or neural network and uses insights from this analysis to formulate a precise conjecture. Next, the agent searches for counterexamples using mathematical software. If they are found, they are used to refine the conjecture and provide additional data for the next small model training and evaluation run. At any point, the agent can attempt to prove the conjecture.

\textbf{Conjectures that PatternMiner proposed and proved:} We applied PatternMiner to five problems from Richard Stanley’s famous survey of positivity problems in algebraic combinatorics \citep{stanley2000}. We chose to focus on this set of problems because (i) the problems are widely accepted as being of high interest to the community, (ii) most questions are broad enough that there are many ways to sharpen them into more specific conjectures, and (iii) algebraic combinatorics tends to be more accessible than some other areas of mathematics like algebraic geometry or pure representation theory where it would be hard to even outline the problem for a general audience. 

PatternMiner formulated and then proved conjectures related to four of these problems. For Problem~11, which involves finding a combinatorial interpretation of Schubert polynomial structure constants, PatternMiner gave a recursive construction showing that projection tests retaining a fixed number of flag dimensions cannot detect every vanishing structure constant (Section~\ref{sec:schubert}; Theorem~\ref{thm:schubert-projection}). It also proves a new equivariant positivity inequality that resolves Conjectures A and B of \citet{chen2026_support} on nonvanishing of equivariant Schubert structure constants. For Problem~12, PatternMiner derived a generating-function formula and an efficient algorithm for character-table row sums (Theorem~\ref{thm:row-evaluation}). For Problem~16, it proved restrictions on a possible failure of Kazhdan--Lusztig combinatorial invariance at length 13 (Theorem~\ref{thm:kl-twelve}). After this work was submitted, this result was subsumed by the resolution of this conjecture by OpenAI \cite{OAI:Combinatorial-Invariance-of-Kazhdan-Lusztig-Polynomials-September-24-2026}. For Problem~18, which involves Schur positivity of monomial immanants, PatternMiner extends the known nonnegativity of Schur coefficients indexed by hook partitions \citep{pmLesnevich2024,pmHikita} to partitions in which the length of every row below the first row is at most two. We provide a more detailed description of PatternMiner's results on these and other problems in Section~\ref{sec:experiments} along with a summary in Table~\ref{tab:summary}.
% \hjk{[there is a word missing here]}

Across the five studies, agents commonly shifted questions, refined conjectures using counterexamples, and restricted difficult problems to more tractable subclasses. Direct-agent baselines produced useful but limited results, while a controlled ablation found that mechanistic intervention evidence improved both model-component selection and the generalization of the resulting mathematical rule.

In summary, our contributions in this work are as follows.
\begin{itemize}[leftmargin=1em]
    \item We propose the PatternMiner framework for conjecture generation and resolution which combines LLM theory-based reasoning with small model, example-driven exploration through automated mechanistic interpretability (Section~\ref{sec:method}).
    \item We describe new mathematical results that come from PatternMiner, with a focus on positivity problems in algebraic combinatorics.
    \item We perform a cross-problem analysis of the successes and failures of this system, helping to better characterize what settings PatternMiner performs best in (Section~\ref{sec:discussion}).
\end{itemize}

\section{Background and related work}
\label{sec:background}

\paragraph{Learning and conjecture generation}

Many conjectures begin as patterns observed in computed examples. Classical automated conjecture generation searches for relations among prescribed numerical invariants \citep{larson2016, fajtlowicz1988conjectures}. 
\citet{davies2021} trained predictors and used gradient attribution to discover patterns that led to a theorem relating knot invariants and a conjecture about Kazhdan--Lusztig polynomials. PatternMiner automates the interpretation step. Small models have also been used to study quiver mutation and predict intersection numbers \citep{he2025,hashemi2025}, and to study problems in geometry and topology \citep{carifio2017machine,gukov2024rigor}. The Algebraic Combinatorics Dataset Repository provides nine tasks for conjecture generation from partitions, permutations, and partial orders \citep{chau2025}.

PatternMiner utilizes LLM agents to perform several steps including data generation and model training, model analysis and interpretation, and conjecture refinement. 
\citet{wagner2021} used the cross-entropy method to refute combinatorial conjectures. PatternBoost alternates local search with transformer training \citep{patternboost2024}. FunSearch and AlphaEvolve evolve LLM-generated programs, obtaining new cap-set constructions and improved algorithms \citep{funsearch2024,alphaevolve2025}.
Language models can also propose conjectures and proof steps. 
Neural proof search constructs derivations of supplied statements \citep{lample2022}, and AlphaGeometry combines language-model proposals with symbolic deduction \citep{trinh2024}. MECA addresses conjecture generation by revising candidate statements together with supporting arguments drawn from mathematical literature \citep{meca2026}. 
MECA organizes this search around mechanisms: principles, such as invariants or reductions, that suggest hypotheses. 
PatternMiner differs from these prior examples in that it utilizes small model training and interpretability in addition to reasoning to suggest and refine conjectures.

\paragraph{Reverse engineering mathematics with interpretability tools}
Interpretations of small models have revealed mathematical structure in their computations and in some cases yielded explicit algorithms. Analyses of modular addition and multiplication for finite-groups and monoids have recovered computations using Fourier features and group representations \citep{nanda2023,chughtai2023, chen2026_multiplication}. While in \citet{kvinge2026can}, the authors more generally looked at whether small neural networks learn concepts, like the commutativity of a group, while learning to solve a narrow task. \citet{huang2026} interpreted a small transformer, derived a traversal algorithm on Dyck paths and proved its equivalence to the zeta map under their labeling convention. PatternMiner utilizes this as a new approach to mathematical discovery via LLMs.

\section{PatternMiner}
\label{sec:method}

At a high level, PatternMiner is an iterative cycle which involves formulating a surrogate task appropriate for a small model, generating training data, training the small model, interpretation and analysis of this model, conjecture refinement and literature comparison, and counterexample mining (Figure~\ref{fig:pipeline}).
The surrogate task, training data, and conjecture may change during exploration. The process continues until its resource budget is exhausted or the user ends the investigation, and it is therefore limited to problems for which many examples can be computed.

PatternMiner uses a lightweight markdown-based workflow, compatible with a generic coding agent, containing five prompts, four record templates, and three method guides for experiment design, model training, interpretation, and conjecture development. The record templates record the project state across sessions and document human steering.

\begin{itemize}[leftmargin=1em]

\item \textbf{Mathematical objective and other initial input:} The input to PatternMiner is a mathematical objective. An example that we describe below is: “Seek structural criteria, formulas, and constructions for Schubert structure constants, particularly their vanishing.” Other input includes a computational budget, a time budget, and suggested literature.

\item \textbf{Formulating small model surrogate tasks:} To pursue the objective, at each PatternMiner cycle the LLM agent selects a predictive task that it can train a small model to solve. Included in task design is the way in which to represent the mathematical objects, a choice that can have significant impacts on what the model learns \citep{scullen2025permutations}.

\item \textbf{Generating training and evaluation data:} The agent assembles a training set. If this is the first iteration of the cycle, this means generating all the data using mathematical software like SageMath \citep{sagemath}. If it is not the first iteration, the agent may mix existing data from a previous iteration with newly discovered counterexamples or other data that is relevant to the current conjecture. This data is split into train and test sets. Examples that were not directly generated by mathematical software are always verified by this software. 

\item \textbf{Training and evaluating small models:} The agent chooses a small lightweight machine learning model to fit to the current training set. This could be a simple MLP, a small transformer, gradient-boosted trees, or even a logistic regression model. 

After training, the model is evaluated on disjoint test data. This data is organized into different populations such as uniformly sampled examples, counterexamples mined in a previous iteration of PatternMiner, or out-of-distribution examples. Performance is compared against a naive baseline such as always selecting the largest class in a classification task.

\item \textbf{Small model interpretation:} For trees and linear models, the agent examines split thresholds, feature importance, and fitted coefficients to identify mathematical quantities associated with the predictions. For neural networks, it compares inputs that produce similar activations at a chosen neuron or layer, proposes a mathematical property they share, and fits a probe to predict that property from the activations.

The agent verifies interpretability hypotheses on unseen data, raw data (in the case of probes), and untrained networks. Other causal tests include activation patching \citep{conmy2023,zhang2024}. 

\item \textbf{Comparison to the literature:} Sometimes a proposed refinement to a conjecture is already supported by an existing result. In other cases, the conjecture may actually be vacuous. At this step, the agent compares the conjecture to the existing literature and considers its consequences. Specifically, the agent looks for the closest known results, asks whether the conjecture is an immediate consequence of these, and evaluates whether its resolution would clarify, organize, or rule out active areas of inquiry. Importantly, when the agent uses a result from the literature, it also checks it against counterexamples in software.

\item \textbf{Search for examples that test and stretch a conjecture:} Given a candidate conjecture, the agent directs a targeted software-driven search for examples that minimally break the conjecture or stretch the conditions of the conjecture. In the case of characterization problems ‘Give conditions for $A$ to be true’, the latter might include examples which satisfy necessary conditions but for which the property does not hold. It might also include examples where sufficient conditions are not satisfied, but the property holds. 

\item \textbf{Conjecture proofs:} At any point during the cycle, the agent can attempt to prove the conjecture or a special case of the conjecture. To help remove incorrect proofs, all proofs receive an adversarial review. 
\end{itemize}

Given a mathematical objective and resource limits, PatternMiner runs as an autonomous process. However, since discoveries are ultimately for the humans overseeing the system, PatternMiner was designed to give legibility into its progress during exploration. This is primarily accomplished via the records that the system generates. This also allows for easy human intervention into the system in cases where the model has chosen a trajectory that is of less interest to the user. 

\section{Experiments}
\label{sec:experiments}

We apply PatternMiner to five problems from Stanley's survey of positivity problems in algebraic combinatorics \citep[Problems~10, 11, 12, 16, and~18]{stanley2000}. A central theme in algebraic combinatorics is the observation that many phenomena from abstract algebra can be re-interpreted in terms of combinatorics. As a simple example, the coefficient on the term $x^k$ in the expansion $(x+1)^n$ counts the number of ways to select $k$ distinct elements from a set of $n$ elements. Such an interpretation enables mathematicians to translate a purely algebraic problem into a problem in combinatorics, unlocking new insights and problem solving techniques. 

We group our discussion below around the five Stanley problems. Four out of the five investigations produced novel results with no counterparts in the literature (to the best of our knowledge). Table~\ref{tab:summary} summarizes our results and Appendix~\ref{app:details} records the data selection, models, controls, checks, and proofs behind each. Note that because of the structure of PatternMiner, results for a given mathematical objective need not follow a single path of inquiry. Indeed, as the descriptions below show, in several cases PatternMiner achieved results from several angles.

\begin{table}[htbp]
\centering\small
\setlength{\tabcolsep}{4pt}
\caption{Mathematical outcomes and research sessions by problem. Generated conjectures include statements subsequently proved or refuted.}
\label{tab:summary}
\begin{tabularx}{\linewidth}{@{}>{\raggedright\arraybackslash}p{.13\linewidth}rY>{\raggedright\arraybackslash}p{.19\linewidth}>{\raggedright\arraybackslash}p{.22\linewidth}@{}}
\toprule
Problem & Sessions & New results & Rediscovered results & Generated conjectures\\
\midrule
Schubert (11) & 61 &
Projection obstruction and equivariant positivity (Thms~\ref{thm:schubert-projection}, \ref{thm:schubert-positivity}); insufficient candidate statistics (Prop.~\ref{prop:schubert-candidate-data}) &
Descent and essential-rank conditions (Props~\ref{prop:schubert-descent}, \ref{prop:schubert-rank}) &
Proved: arbitrarily large projection obstructions (Thm.~\ref{thm:schubert-projection})\\
\midrule
Kronecker (10) & 17 & --- &
First-row inequalities and hook bounds (Thms~\ref{thm:kronecker-murnaghan}, \ref{thm:kronecker-rosas}) &
Refuted: two-row triangle inequalities (Prop.~\ref{prop:kronecker-truncation}); empirical: distinct-parts ordering (Table~\ref{tab:kronecker-comparison})\\
\midrule
Row sums (12) & 33 &
Fixed-tail formula and evaluation algorithm (Thm.~\ref{thm:row-evaluation}) &
Positive formulas for hook partitions (App.~\ref{app:recovered-row-formulas}) &
Refuted: injectivity of learned tableau map (Table~\ref{tab:row-checks})\\
\midrule
Kazhdan--Lusztig (16) & 62 &
Restrictions on length-13 failures (Prop.~\ref{prop:kl-difference}; Thm.~\ref{thm:kl-twelve}; Prop.~\ref{prop:kl-descent-reduction}) &
Invariance through length 12 (Thm.~\ref{thm:kl-known}) &
Proposed: two coefficient bounds (Conjs~\ref{conj:kl-descent}, \ref{conj:kl-genus-zero})\\
\midrule
Monomial immanants (18) & 32 &
Schur positivity for $\theta_2\leq2$ (Thm.~\ref{thm:immanant-positivity}) & --- &
Proved: nonnegative character decomposition (Thm.~\ref{thm:immanant-positivity})\\
\bottomrule
\end{tabularx}
\end{table}

\subsection{Schubert structure constants (Stanley's Problem 11)}
\label{sec:schubert}

\paragraph{Background} A \emph{flag} is a nested sequence of subspaces of a fixed vector space. For permutations $u,v,w$, the \emph{Schubert structure constant} $c_{u,v}^w$ is a nonnegative integer that counts flags satisfying incidence conditions. Equivalently, these integers are the coefficients in the expansion of a product of \emph{Schubert polynomials}, which are polynomial representatives of cohomology classes of Schubert varieties, indexed by permutations:
\[\mathfrak{S}_u \cdot \mathfrak{S}_v = \sum_w c_{u,v}^w \mathfrak S_w.\]

\begin{figure}[!htbp]
\centering
\includegraphics[width=.93\linewidth]{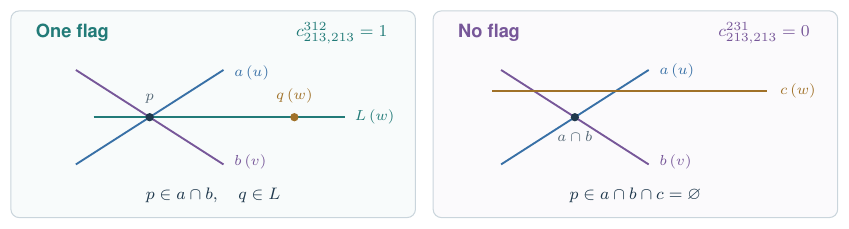}
\caption{\textbf{Schubert structure constants count flags satisfying incidence conditions.} In both panels $u=v=213$ fixes $p=a\cap b$. For $w=312$ the line $L$ passes through $p$ and $q$. For $w=231$ the point $p$ must also lie on $c$.}
\label{fig:schubert}
\end{figure}

Finding a combinatorial interpretation of these numbers is a central open problem in Schubert calculus. Figure~\ref{fig:schubert} illustrates a coefficient equal to one and a coefficient equal to zero. We study two weaker questions: when do these coefficients vanish, and which operations on permutations preserve nonvanishing? For a comprehensive introduction, we refer the reader to \citet{billey2025}.

\paragraph{Results} A projection test retains selected flag dimensions and asks whether the projected Schubert incidence conditions have a common solution; an empty intersection certifies vanishing.
\begin{itemize}[leftmargin=1em]
    \item Transformers trained on Schubert products suggested the classical descent condition: $c_{u,v}^{w}>0$ implies $\Des(w)\subseteq\Des(u)\cup\Des(v)$, where $\Des(x)=\{i:x(i)>x(i+1)\}$ \citep{knutson2003_recurrence}.
    \item Rank-matrix feature searches on triples passing the existing tests suggested a second classical condition: if the three \emph{permutation rank matrices} agree at an \emph{essential position} of $w$, then $c_{u,v}^{w}=0$. This is a consequence of the Littlewood--Richardson rule \citep{fulton1997}; see Proposition~\ref{prop:schubert-rank}.
    \item Hard-negative mining led to a recursive construction: for every $n\geq5$, there exist $u,v,w\in S_n$ with $c_{u,v}^{w}=0$ and $|\Des(u)\cup\Des(v)|=n-3$, yet every projection to a \emph{partial flag variety} retaining a proper subset of these dimensions has a nonempty generic intersection. Thus no fixed number of retained dimensions suffices for these projection tests (Theorem~\ref{thm:schubert-projection}).
    \item Computations suggested a positivity inequality for the \emph{equivariant structure constants} $c_{u,v}^{w}(t)$, whose constant terms are the ordinary structure constants. We prove
    \begin{equation}
    c_{u,v}^{w}(t)-\alpha_i c_{s_i u,v}^{w}(t)
    \in\Z_{\geq0}[\alpha_1,\ldots,\alpha_{n-1}]
    \qquad\text{if }s_i u<u\leq w,
    \label{eq:schubert-main-inequality}
    \end{equation}
    where $\alpha_i=t_{i+1}-t_i$, $s_i$ swaps the values $i,i+1$, and $\leq$ is \emph{Bruhat order}. Nonvanishing is therefore preserved when the first factor increases in \emph{left weak order} below $w$. This proves Conjectures A and B of \citet{chen2026_support} on equivariant nonvanishing (Theorem~\ref{thm:schubert-positivity} and Corollary~\ref{cor:schubert-support}).
     \item A six-layer transformer learned to select $j$ with $c_{u,s_jv}^{w}>0$ given $c_{s_i u,v}^{w}>0$, under the hypotheses in Appendix~\ref{app:schubert-models}, where Corollary~\ref{cor:schubert-support} guarantees a valid $j$. The agent refined its training data by mining prediction errors. Activation patching indicates that the model relies partly on the positions of values outside the candidate pairs $(j,j+1)$ to choose its answer; Appendix~\ref{app:schubert-selector} reports the measurements and controls.
\end{itemize}

\paragraph{Discussion} The agent needed human redirection at one decisive point. In the projection tests above, the agent repeatedly sought zeros requiring five, six, and then seven retained flag dimensions for the intersection of the projected Schubert varieties to be empty. We directed it toward a construction valid for every $n\geq5$; Theorem~\ref{thm:schubert-projection} shows no fixed number of retained dimensions suffices. The agent also trained a classifier whose output depended on the order of $u$ and $v$, although $c_{u,v}^w=c_{v,u}^w$. Representation choices also mattered: sparse-autoencoder features gave AUC $0.9854$, compared with $0.9855$ for random linear projections of the same activations (Table~\ref{tab:schubert-features}).

\subsection{Kronecker coefficients (Stanley's Problem 10)}
\label{sec:kronecker}

\begin{figure}[!htbp]
\centering
\includegraphics[width=.93\linewidth]{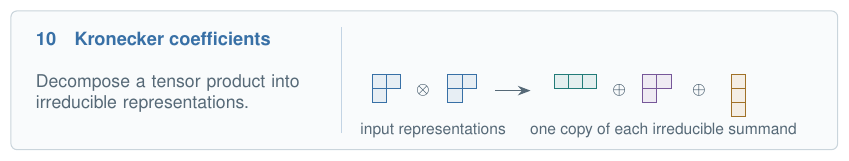}
\caption{\textbf{Kronecker coefficients give tensor-product multiplicities.}}
\label{fig:kronecker}
\end{figure}

\paragraph{Background} The \emph{irreducible representations} of the \emph{symmetric group} $S_n$ are indexed by \emph{partitions} $\lambda\vdash n$. The \emph{Kronecker coefficient} $g_{\lambda,\mu,\nu}$ is the nonnegative integer multiplicity of $V^\nu$ in the \emph{tensor product} $V^\lambda\otimes V^\mu$. Stanley's Problem~10 asks for a combinatorial interpretation of these multiplicities. Figure~\ref{fig:kronecker} shows a tensor product for $S_3$ in which each of three irreducible representations occurs once.

\paragraph{Results}
\begin{itemize}[leftmargin=1em]
    \item Gradient-boosted trees trained to predict vanishing concentrated feature importance on the largest parts, recovering Murnaghan's necessary condition: $g_{\lambda,\mu,\nu}>0$ implies $n-\lambda_1,n-\mu_1,n-\nu_1$ satisfy the triangle inequalities \citep[Equation~(22)]{mishna2021}. Counterexamples ruled out the analogous condition on the first two rows.
    \item Hooks were overrepresented among correctly predicted zeros satisfying Murnaghan's inequalities. Follow-up classifiers for $g\leq1$ and $g\leq2$ led the agent to Rosas's bound $g_{\lambda,\mu,\nu}\leq2$ when two partitions are hooks \citep[Corollary~15]{rosas2001}. A later classifier respected this bound on all 5,236 applicable test triples.
    \item The total number of distinct parts across the three partitions predicts relative coefficient size, with reversal rates of $0.44\%$ ($n=10,11$) and $1.41\%$ ($n=15$) when totals differ by at least four (Table~\ref{tab:kronecker-comparison}). At $n=10,11$, its Spearman correlation with coefficient size is $0.6421$, compared with $0.4857$ for the total \emph{Durfee-square size}, which controls growth in Pak--Panova's bounds \citep{pakpanova2023} (Table~\ref{tab:kronecker-correlation}).
\end{itemize}

\paragraph{Discussion} The investigation recovered classical bounds and identified an empirical relation between the number of distinct parts and coefficient size. Appendix~\ref{app:kronecker} gives sampling conditions and the full experimental results.

\subsection{Character-table row sums (Stanley's Problem 12)}
\label{sec:characters}

\begin{figure}[!htbp]
\centering
\includegraphics[width=.93\linewidth]{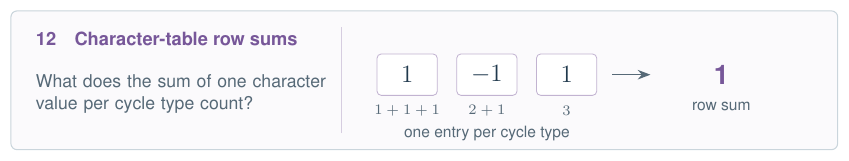}
\caption{\textbf{Character-table row sums combine positive and negative entries.}}
\label{fig:characters}
\end{figure}

\paragraph{Background} The \emph{character table} of a finite group encodes much of its representation theory. For $S_n$, the entry $\chi^\lambda(\mu)$ is the irreducible character indexed by $\lambda\vdash n$, evaluated on permutations of cycle type $\mu\vdash n$. The \emph{character-table row sums} $R(\lambda)=\sum_{\mu\vdash n}\chi^\lambda(\mu)$ are nonnegative even though individual entries can be negative. Stanley's Problem~12 asks for a combinatorial interpretation of these numbers. Column sums, by contrast, have a classical interpretation: the column sum at $g\in S_n$ counts the $\sigma\in S_n$ with $\sigma^2=g$ \citep{sagan2001}.
Figure~\ref{fig:characters} shows the sign-character row of $S_3$, whose entries $1,-1,1$ sum to one.

\paragraph{Results}
\begin{itemize}[leftmargin=1em]
\item We prove that, for partitions $(n-k,\nu)$ with fixed $\nu\vdash k$, the row-sum generating series is the partition generating function times a rational function. Using the stable character polynomials of \citet{garsia2009}, the resulting exact evaluation algorithm takes $\exp(O(\sqrt{k}))\,\mathrm{poly}(n)$ bit operations. Thus row sums can be evaluated in polynomial time when $k=O((\log n)^2)$, whereas direct summation uses $p(n)=\exp(\Theta(\sqrt n))$ conjugacy classes. Appendix~\ref{app:row-evaluation} gives the formula and complexity argument.
    \item A linear model trained to pair negative tableaux with positive tableaux suggested a sign-reversing injection, but collisions on held-out partitions refuted injectivity (Appendix~\ref{app:row-learning}).

\end{itemize}

\paragraph{Discussion}
We found that the hardest part was turning a rule learned
from finite data into a uniform combinatorial construction.
Row-sum calculations did not reveal what the sums count.
To make further progress, we need a uniform cancellation
rule whose injectivity can be proved beyond the observed data.

\subsection{Kazhdan--Lusztig coefficients (Stanley's Problem 16)}

\begin{figure}[!htbp]
\centering
\includegraphics[width=.93\linewidth]{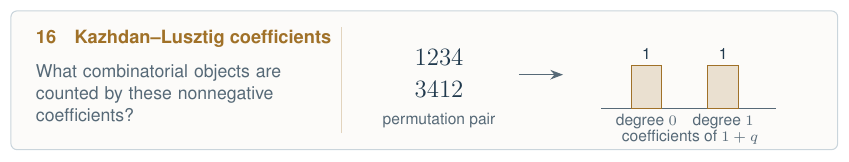}
\caption{\textbf{Kazhdan--Lusztig polynomials have nonnegative coefficients.}}
\label{fig:kazhdan-lusztig}
\end{figure}

\paragraph{Background} \emph{Kazhdan--Lusztig polynomials} $P_{u,w}(q)$ are indexed by pairs of permutations $u\leq w$ in \emph{Bruhat order} and have nonnegative integer coefficients. For example, $P_{1234,3412}(q)=1+q$ (Figure~\ref{fig:kazhdan-lusztig}). Stanley's Problem~16 asks for a combinatorial interpretation of these coefficients. We study the related conjecture of \emph{combinatorial invariance}, due to Lusztig and Dyer: whether the poset isomorphism type of $[u,w]$ determines $P_{u,w}$ \citep{dyer1987,blundell2022}. Previous work used neural prediction and saliency analysis to guide progress on this conjecture \citep{davies2021}. The interval length is $\ell(w)-\ell(u)$, where $\ell$ counts inversions. At the time of the investigation, invariance was known for type-A intervals of length at most 12 \citep[Corollary~1.3]{pmEMS2026}, and we focused on length 13. The subsequent resolution of the entire conjecture by OpenAI is cited in the introduction.

\paragraph{Results}
\begin{itemize}[leftmargin=1em]
    \item Decision stumps were trained to classify individual recurrence terms as positive or zero. The two best stumps (tied, 30/33) both failed on a withheld pair, ruling out the selected single-feature thresholds (Appendix~\ref{app:kl-learning}).
    \item We prove that a failure of combinatorial invariance at length 13 would require an interval $[u,w]$ in $S_{12}$ such that $u^{-1}w$ has exactly three cycles and $[q^6]P_{u,w}\geq126$. The argument combines a relation among the possible coefficient differences, cancellation of common permutation entries, and Warrington's computations through $S_{11}$ \citep{pmWarrington}; see Theorem~\ref{thm:kl-twelve}.
    \item Iterating the Kazhdan--Lusztig recurrence shows that such a failure would force the coefficient of $q^6$ to decrease by at least five at one of at most 26 common-descent steps (Proposition~\ref{prop:kl-descent-reduction}). A bound of four for every decrease in the specified class would therefore establish invariance at length 13 (Conjecture~\ref{conj:kl-descent}).
\end{itemize}

\paragraph{Discussion} Counterexamples directed the investigation from predicting individual terms toward controlling their combined contribution in the recurrence. The failed thresholds exposed a limitation of the selected features. A later example also ruled out a simpler coefficient bound that omitted the recurrence's correction terms (Table~\ref{tab:kl-counterexample}). The resulting reductions isolate an inequality for further study; they follow from computations and algebraic arguments. Appendix~\ref{app:kazhdan-lusztig} gives the proofs and the remaining conjectures.

\subsection{Monomial immanants (Stanley's Problem 18)}

\begin{figure}[!htbp]
\centering
\includegraphics[width=.93\linewidth]{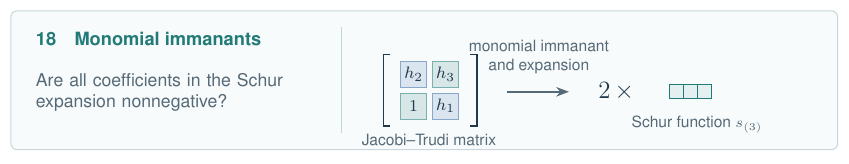}
\caption{\textbf{Monomial immanants are expanded in Schur functions.}}
\label{fig:immanants}
\end{figure}

\paragraph{Background} An \emph{immanant} generalizes the determinant by replacing its signs with other permutation weights. Stanley's Problem~18 asks whether \emph{monomial immanants} of \emph{Jacobi--Trudi matrices} $H_{\mu/\nu}$ are \emph{Schur-positive}: whether their expansions in Schur symmetric functions $s_\theta$ have nonnegative coefficients. These matrices are indexed by skew shapes $\mu/\nu$; Figure~\ref{fig:immanants} gives an example whose expansion is $2s_{(3)}$. For hook partitions $\theta$, whose diagrams consist of one row and one column meeting at a corner, Lesnevich decomposed the associated immanant character into a nonnegative integer combination of Stanley--Stembridge characters \citep{pmLesnevich2024}. Hikita's positivity theorem for these characters implies nonnegativity of the corresponding Schur coefficients \citep{pmHikita}. We investigate how far this decomposition extends beyond hooks.

\paragraph{Results}
\begin{itemize}[leftmargin=1em]
    \item We prove that the coefficient of $s_\theta$ is nonnegative in every monomial immanant of $H_{\mu/\nu}$ whenever $\theta_2\leq2$. This permits arbitrarily many rows of length two below the first row and places no restriction on the skew shape. The proof decomposes the associated immanant character into a nonnegative integer combination of \emph{Stanley--Stembridge characters}; Hikita's theorem supplies positivity for each summand \citep{pmHikita}. Theorem~\ref{thm:immanant-positivity} gives the statement.
    \item Computations verified nonnegativity of 776,891 decomposition coefficients across 1,587 shape instances, together with the auxiliary identities used in the proof. Appendix~\ref{app:immanant-checks} records the finite checks and their conventions.
\end{itemize}

\paragraph{Discussion} The key step was to express the decomposition through \emph{Kostka numbers}, which count tableaux of prescribed shape and content. Extending the hook calculation led to a generating function for coefficient partitions with every row below the first of length at most two. Factoring the resulting alternating sums then gave a proof for the entire family. The finite computations checked the formulas and searched for negative coefficients; the general positivity result rests on the symbolic factorization in Appendix~\ref{app:immanants}. A later classifier tested whether character-data vectors came from an enumerated collection of skew shapes (Appendix~\ref{app:cross-study-controls}).

\section{Analysis of the discovery process}
\label{sec:discussion}

In this section, we examine common conjecture generation strategies, system failure modes, and comparison to direct LLM baselines.

Having analyzed the logs from the case studies above, we identified the following as common patterns in conjecture generation.
\begin{itemize}[leftmargin=1em]
\item \textbf{Turning feature associations into candidate conjecture conditions:} In the Kronecker study, feature importance in a small model highlighted the largest partition parts, leading to Murnaghan's first-row inequalities \citep{mishna2021}.
\item \textbf{Use unexplained examples to motivate new conjectures or conjecture refinement:} It was common for the agent to search for examples that satisfy every necessary condition (for a known theorem or the candidate conjecture) but fail the property. Analogously, the agent also searched for examples that fail every sufficient condition, but satisfy the property. For example, in the Schubert study, the agent searched for zero-valued structure constants that were undetected by every parabolic projection test retaining a proper subset of the descent dimensions.
\item \textbf{Refining a conjecture based on counterexamples:} Here, the agent looked for a common feature among the examples that a conjecture left unexplained. Zeros satisfying Murnaghan's inequalities prompted a two-row extension, which a counterexample refuted (Proposition~\ref{prop:kronecker-truncation}). Hooks among the remaining zeros then suggested a more specific family to study (Theorem~\ref{thm:kronecker-rosas}).
\item \textbf{Question shifting:} In the Kronecker study, the agent shifted from vanishing to small-value thresholds and then to comparisons between coefficients in an effort to find a fruitful direction of study.
\item \textbf{Restricting to a smaller class of objects:} In the immanant study, the correction to the known hook formula had a negative coefficient. The hook contribution canceled it. Proving this cancellation settled the first non-hook case and led to the theorem for all coefficient partitions with $\theta_2\leq2$ (Appendix~\ref{app:immanant-conjecture-history}).
\end{itemize}

\begin{table}[!htbp]
\centering
\caption{Counts of observed discovery patterns across the five problems.}
\label{tab:pattern-counts}
\small
\setlength{\tabcolsep}{4pt}
\begin{tabular}{@{}lrrrrrr@{}}
\toprule
Pattern
  & \shortstack{Kronecker}
  & \shortstack{Schubert}
  & \shortstack{Character\\tables}
  & \shortstack{Kazhdan--\\Lusztig}
  & \shortstack{Imma-\\nants}
  & Total \\
\midrule
Feature associations       & 3 & 4 & 2 & 1 & 1 & \textbf{11} \\
Unexplained examples       & 3 & 9 & 0 & 1 & 0 & \textbf{13} \\
Counterexample refinement  & 4 & 4 & 3 & 1 & 1 & \textbf{13} \\
Question shifting          & 5 & 9 & 2 & 1 & 0 & \textbf{17} \\
Restriction to a subclass  & 1 & 5 & 4 & 2 & 1 & \textbf{13} \\
\bottomrule
\end{tabular}
\end{table}

We used Opus 5.5 \citep{anthropic2026opus55} to identify instances of these patterns across the five study records. Table~\ref{tab:pattern-counts} shows that question shifting was the most common pattern, although its frequency varied substantially across problems.

\subsection{Direct LLM baselines.}
Does the PatternMiner workflow really result in different behavior than one would achieve by simply using an LLM without the harness? To test this, we gave each of two model configurations (GPT-6.2 Astra and GPT-6.2 Sol) one 15-minute attempt on each of the five Stanley problems, without the PatternMiner workflow or a trained predictor. The direct agents often found explicit formulas for special cases. In comparison, PatternMiner tended to find more general conjectures (and in some cases, theorems). On the Schubert problem, the direct agents found rules for computing coefficients on particular families of inputs. PatternMiner, on the other hand, investigated examples where a proposed test missed a zero coefficient and proved that, for any fixed number of dimensions the test retains, there are zero coefficients it will miss. It’s possible that by grounding an LLM in an experimental workflow, with continual feedback from examples and model interpretations results in better research taste. 

Of course, without equivalent compute budgets, its impossible to make fair and rigorous direct comparisons. We hope to leave this comparison for the future. The interested reader can consult Appendix~\ref{app:direct-llm-baselines} for the direct LLM results on all five problems, two supplemental retries, and a separate matched Schubert comparison.

\subsection{Model-interpretation ablation.}
\label{sec:ablations}
We wanted to know whether looking inside a trained model helps an agent write a better mathematical rule than looking at its predictions alone. To test this, we gave two agents the same predictions from a trained Schubert model. Only one agent also saw how the predictions changed when we replaced an attention head’s output on one example with its output on the other example in the pair. We asked each agent to write a rule for choosing which adjacent values to swap in the second permutation so that the resulting Schubert structure constant is positive. We tested both rules on the same pairs of examples that neither agent had seen. The rule based on predictions alone chose a valid swap for both examples in 174 of the 249 unseen pairs. The rule that also used the attention-head results did so in 214 pairs. This suggests that looking inside the model can help even when its predictions alone are enough to write a useful rule. Appendix~\ref{app:interpretation-ablation} gives the full setup and results for this comparison.

\subsection{Hard-negative-mining ablation.}
Without hard-negative mining during training, the Schubert investigation found six vanishing examples that survived descent cycling but did not reach the general construction obtained by the full pipeline. Appendix~\ref{app:hard-negative-ablation} reports this comparison and another matched sampling comparison.

\subsection{Failure modes}

\begin{itemize}[leftmargin=1em]
\item \textbf{Collecting larger and larger sets of examples with corresponding added conditions without arriving at a result:} The Schubert search repeatedly produced zeros requiring more retained flag dimensions. We redirected it toward a construction valid for every $n\geq5$ (Theorem~\ref{thm:schubert-projection}). This search illustrates the repetitive action looping described by \citet{chen2026deloop}.
\item \textbf{Incomplete comprehension of the literature:} In several cases, agents extracted information from a paper without capturing all the relevant conditions. In math, conventions are often established at the beginning of a section, far from the actual statement. This led to agents utilizing an existing theorem in an incorrect way. This occurred when a PatternMiner agent tried to use a theorem related to Kronecker coefficients from \citep{mishna2021}.
\item \textbf{Interpretations that add nothing new:} In some cases, model interpretability yielded patterns that could not be resolved into a concise conjecture or simply confirmed facts that were already known. These were challenging to build new conjectures around.
\end{itemize}

\paragraph{Limitations}
\label{sec:limitations}
Our experiments are limited to algebraic combinatorics. We have not compared PatternMiner directly with FunSearch or AlphaEvolve, even though PatternMiner provides a different mode of mathematical discovery from the other two.
The direct agents and PatternMiner started from the same broad mathematical questions, but their tools, available context, human guidance, and research budgets differed. Each baseline configuration received one initial attempt per problem, and agents chose their own subquestions.

The number of sessions for the Schubert structural constant problem in Table~\ref{tab:summary} was reconstructed by question and recorded outcome, while the other counts came from experiment logs, limiting direct comparison across problems. Each LLM run is expensive and difficult to replicate, which limits our ability to conduct a fully controlled study.

\bibliography{references}
\bibliographystyle{tmlr}

\appendix
\numberwithin{equation}{section}
\raggedbottom
\setlength{\intextsep}{8pt plus 2pt minus 2pt}
\setlength{\abovedisplayskip}{5pt plus 2pt minus 3pt}
\setlength{\belowdisplayskip}{5pt plus 2pt minus 3pt}

% adding a section for condensing all the stanley problem statements in one spot
\section{Problem statements}
\label{app:problems}

We state the five questions from \citet{stanley2000} in the notation of this paper, followed by the tasks pursued in our investigations. A combinatorial interpretation means an explicit family of objects counted by the quantity in question. Its nonnegativity or numerical evaluation alone does not supply that interpretation.

\paragraph{Problem 10: Kronecker coefficients}
For partitions $\lambda,\mu,\nu\vdash n$, define $g_{\lambda,\mu,\nu}$ by the decomposition of irreducible characters
\[
 \chi^\lambda\chi^\mu=\sum_{\nu\vdash n}g_{\lambda,\mu,\nu}\chi^\nu.
\]
The problem asks for a combinatorial interpretation of these coefficients. Our prediction tasks concern vanishing, small values and magnitude; recovered bounds and empirical comparisons do not give a general counting rule.

\paragraph{Problem 11: Schubert structure constants}
For Schubert polynomials indexed by permutations with finite support, write
\[
 \mathfrak S_u\mathfrak S_v=\sum_w c_{u,v}^w\mathfrak S_w.
\]
The problem asks for a combinatorial interpretation of $c_{u,v}^w$. We study necessary conditions for nonvanishing, limits of projection tests, and operations preserving nonvanishing. The learned selector addresses one such operation. These tasks are narrower than constructing a general counting rule.

\paragraph{Problem 12: character-table row sums}
For $\lambda\vdash n$, put
\[
 R(\lambda)=\sum_{\rho\vdash n}\chi^\lambda(\rho).
\]
The sum has one term per conjugacy class, without class-size weights. The problem asks for a combinatorial interpretation of $R(\lambda)$. We derive a fixed-tail evaluation formula and test a learned injection between negative and positive tableaux in a restricted character expansion. That finite injection task does not represent every row sum.

\paragraph{Problem 16: Kazhdan--Lusztig coefficients}
For a finite or affine Weyl group $W$ and $x\leq w$ in Bruhat order, the problem asks for a combinatorial interpretation of each coefficient $[q^k]P_{x,w}(q)$, especially for $W=S_n$. Our investigation instead focuses on the related conjecture of \emph{combinatorial invariance} in type~A:
\[
 [x,w]\cong[x',w']\quad\Longrightarrow\quad P_{x,w}=P_{x',w'}.
\]
Here the isomorphism preserves the partial order, without permutation labels. We learn rules for individual recurrence terms and derive restrictions on a possible length-thirteen failure. Neither invariance nor the original counting problem is resolved in general by this work.

\paragraph{Problem 18: monomial immanants}
Let $D=\mu/\nu$ be a skew shape with at most $n$ rows, and set $H_D=(h_{\mu_i-\nu_j-i+j})_{i,j=1}^n$, where $h_k$ is the complete symmetric function, $h_0=1$ and $h_k=0$ for $k<0$. For $\lambda\vdash n$, let $\phi^\lambda$ be the virtual character with Frobenius characteristic $m_\lambda$. The problem asks whether
\[
 [s_\theta]\operatorname{Imm}_{\phi^\lambda}(H_D)\geq0
 \quad\text{for every }\theta\vdash |D|,\qquad
 \operatorname{Imm}_{\phi^\lambda}(A)=\sum_{w\in S_n}\phi^\lambda(w)\prod_{i=1}^n a_{i,w(i)}.
\]
We prove this positivity for coefficient partitions with $\theta_2\leq2$, through a nonnegative character decomposition. The immanant index $\lambda$, coefficient partition $\theta$ and skew shape $D$ play different roles. The separate classifier experiment tests which character-data vectors occur in an enumerated shape collection; it does not decide the full positivity problem.

The detailed experiments and proofs follow in Appendix~\ref{app:details}.
\clearpage

\section{Mathematical and experimental details}
\label{app:details}

The label \emph{LLM-generated} identifies statements and proofs proposed by an LLM agent during the investigation.

\subsection{Schubert structure constants}
\label{app:schubert}

\subsubsection{Mathematical setting}
Let $\Fl_n(\C)$ be the \emph{complete flag variety}, and let $\sigma_w$ be its \emph{Schubert class} of complex codimension $\ell(w)$, where $\ell$ is the \emph{inversion number}. We use the conventions of \citet{knutson2022}; \citet{fulton1997} gives the background on flag varieties and \emph{Schubert polynomials}. The ordinary structure constants satisfy
\[
 \sigma_u\sigma_v=\sum_{w\in S_n}c_{u,v}^{w}\sigma_w,\qquad c_{u,v}^{w}\in\Z_{\geq0}.
\]
For $\ell(w)=\ell(u)+\ell(v)$, the coefficient counts the intersection of the \emph{Schubert varieties} indexed by $u,v,w_0w$ relative to three general flags. Here $w_0$ is the longest permutation. We write $\Des(w)=\{i:w(i)>w(i+1)\}$ for \emph{right descents} and $\Des_L(w)=\{i:w^{-1}(i)>w^{-1}(i+1)\}$ for \emph{left descents}. Right multiplication by $s_i=(i,i+1)$ swaps positions, and left multiplication swaps values.

The examples in Figure~\ref{fig:schubert} are $c_{213,213}^{312}=1$ and $c_{213,213}^{231}=0$. A flag in $\C^3$ is a point $p$ on a line $L$ in $\mathbb P^2$. Two general conditions on $p$ determine their intersection point, and a condition that $L$ contain another general point determines $L$. Three general conditions on $p$ have no solution. The same counts follow from $\mathfrak S_{213}^2=x_1^2=\mathfrak S_{312}$.

\begin{pipelineoutput}
\begin{figure}[H]
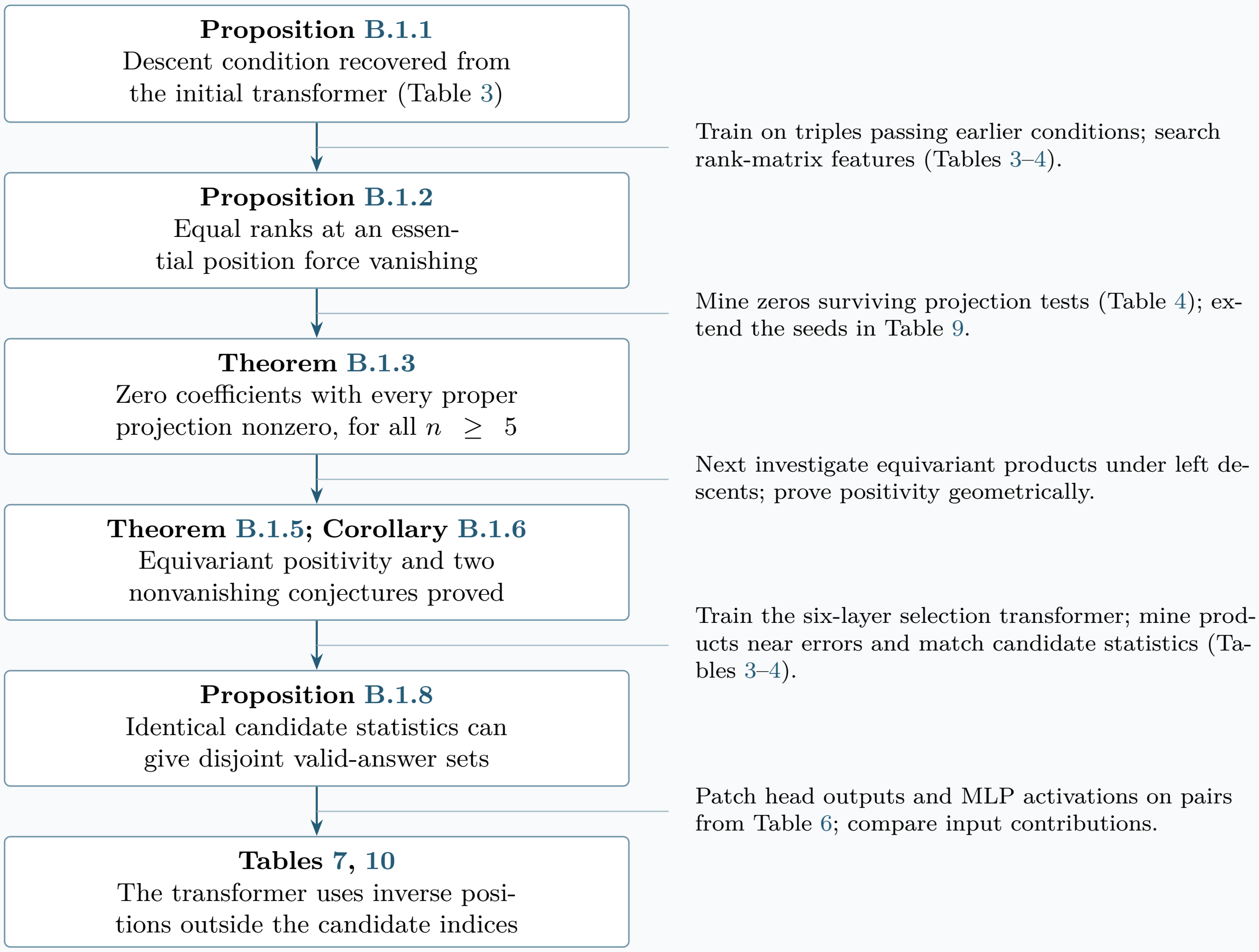

\centering
\begin{discoverychart}
\discoveryfirst{s1}{\textbf{Proposition~\ref{prop:schubert-descent}}\\Descent condition recovered from the initial transformer (Table~\ref{tab:schubert-models})}
\discoverynext{s2}{s1}{\textbf{Proposition~\ref{prop:schubert-rank}}\\Equal ranks at an essential position force vanishing}{Train on triples passing earlier conditions; search rank-matrix features (Tables~\ref{tab:schubert-models}--\ref{tab:schubert-mining}).}
\discoverynext{s3}{s2}{\textbf{Theorem~\ref{thm:schubert-projection}}\\Zero coefficients with every proper projection nonzero, for all $n\geq5$}{Mine zeros surviving projection tests (Table~\ref{tab:schubert-mining}); extend the seeds in Table~\ref{tab:schubert-seeds}.}
\discoverynext{s4}{s3}{\textbf{Theorem~\ref{thm:schubert-positivity}; Corollary~\ref{cor:schubert-support}}\\Equivariant positivity and two nonvanishing conjectures proved}{Next investigate equivariant products under left descents; prove positivity geometrically.}
\discoverynext{s5}{s4}{\textbf{Proposition~\ref{prop:schubert-candidate-data}}\\Identical candidate statistics can give disjoint valid-answer sets}{Train the six-layer selection transformer; mine products near errors and match candidate statistics (Tables~\ref{tab:schubert-models}--\ref{tab:schubert-mining}).}
\discoverynext{s6}{s5}{\textbf{Tables~\ref{tab:schubert-interventions}, \ref{tab:schubert-equal-data}}\\The transformer uses inverse positions outside the candidate indices}{Patch head outputs and MLP activations on pairs from Table~\ref{tab:schubert-pairs}; compare input contributions.}
\end{discoverychart}
\caption{Schubert discoveries and the experiments leading to them.}
\label{fig:app-schubert-flow}
\end{figure}

\subsubsection{Models and hard-negative experiments}
\label{app:schubert-models}
The first task predicts whether $c_{u,v}^{w}>0$ from a degree-compatible triple. A hard negative is a computed zero that passes the necessary conditions used to select its dataset. For the model trained on the remaining triples, those conditions were degree compatibility, $u,v\leq w$ in Bruhat order, the descent condition, and the zero-rank case of Proposition~\ref{prop:schubert-rank}. Permutation products, rather than individual triples from the same product, determine the training and evaluation split.

\begin{table}[H]
\centering\small
\caption{Schubert prediction models.}
\label{tab:schubert-models}
\begin{tabularx}{\linewidth}{@{}>{\raggedright\arraybackslash}p{.22\linewidth}Y>{\raggedright\arraybackslash}p{.28\linewidth}@{}}
\toprule
Task & Model and training data & Recorded evaluation\\
\midrule
$c_{u,v}^{w}>0$ & Transformer: width 192, six heads, four encoder layers and two cross-attention layers; 1,272,767 training triples & Product-disjoint test: F1 $0.992$, Matthews correlation $0.980$\\
$c_{u,v}^{w}>0$ after necessary conditions & Transformer trained on triples passing the conditions above & Matthews correlation $0.961$ on the corresponding evaluation population\\
Select $j$ with $c_{u,s_jv}^{w}>0$ & Transformer: six layers, width 384, sixteen heads per layer; 17,994,585 training examples & Separate $S_{10}$ assessment: $99.799\%$ per supplied input; $1{,}874/1{,}902$ difficult pairs correct\\
\bottomrule
\end{tabularx}
\end{table}

The selection task uses $\ell(w)=\ell(u)+\ell(v)-1$ and a supplied $i\in\Des_L(u)$ with $c_{s_i u,v}^{w}>0$. A correct answer belongs to
\begin{equation}
 \mathcal V(u,v,w)=\{j\in\Des_L(v):c_{u,s_jv}^{w}>0\}.
 \label{eq:witnesses}
\end{equation}
We retain inputs with $\mathcal V(u,v,w)\ne\varnothing$; $u\leq w$ guarantees this by Corollary~\ref{cor:schubert-support}. The final architecture discards the supplied index. Its inputs are inverse positions, descent indicators, rank-matrix entries, and images of simple roots under relative permutations. An auxiliary prediction of $\log(1+c_{u,s_jv}^{w})$ was combined with the selection score into a single linear output head. Coefficient values supply supervision only. Factor labels $S_7$--$S_{10}$ specify the sampled factors; stable products can have output permutations in a larger symmetric group, here of size at most seventeen.

\begin{table}[H]
\centering\small
\caption{Tests used to refine the Schubert tasks.}
\label{tab:schubert-mining}
\begin{tabularx}{\linewidth}{@{}>{\raggedright\arraybackslash}p{.27\linewidth}Y>{\raggedright\arraybackslash}p{.27\linewidth}@{}}
\toprule
Test & Population or selection rule & Result\\
\midrule
Equal-rank condition & Zeros passing the earlier necessary conditions & $133/45{,}000$ excluded\\
Soundness of that condition & Separate nonzero coefficients & $0/59{,}979$ violations\\
Neighboring factor pairs & Products near prediction errors, excluding reserved factor orbits & 4,683,591 new examples and 27,352 difficult pairs in 952 factor groups\\
Proper projections & The stated seeds in $S_5$ and $S_6$ & Two zero coefficients; all eight nonempty proper projections nonzero\\
\bottomrule
\end{tabularx}
\end{table}

The representation comparison in Table~\ref{tab:schubert-features} selects the best feature on the same 59,441 examples. The circuit tests below use a separate confirmation sample.

\begin{table}[H]
\centering\small
\caption{Feature comparison on the same Schubert examples.}
\label{tab:schubert-features}
\begin{tabular}{@{}lr@{}}
\toprule
Representation & Best zero-label AUC\\
\midrule
1,024 sparse-autoencoder features &0.9854\\
1,024 random linear projections &0.9855\\
\bottomrule
\end{tabular}
\end{table}

\subsubsection{Circuit interventions}
\label{app:schubert-selector}
\label{app:interpretation}
Activation patching replaces specified activations on a recipient input with those computed on a donor, then recomputes the affected downstream activations \citep{conmy2023,zhang2024}. For disjoint recipient and donor answer sets $A,B$ and output logits $z$, we measure the change in
\begin{equation}
 \log\!\sum_{j\in B}e^{z_j}-\log\!\sum_{j\in A}e^{z_j}.
 \label{eq:patch}
\end{equation}
A positive change shifts relative probability toward the donor's valid answers. Both directions are evaluated for every pair. To form factor orbits, remove common initial and terminal fixed points, then identify factor pairs under exchange and the simultaneous involution $(u,v)\mapsto(w_0uw_0,w_0vw_0)$. Means weight these orbits equally; bootstrap intervals resample them. Identity replacements and agreement with the original forward computation check the implementation.

\begin{table}[H]
\centering\small
\caption{Samples for the fixed Schubert transformer.}
\label{tab:schubert-pairs}
\begin{tabularx}{\linewidth}{@{}Yrrr@{}}
\toprule
Sample & Pairs & Factor orbits & Both answers correct\\
\midrule
Component selection & 320 & 29 & 316\\
Confirmation & 249 & 22 & 248\\
Earlier pairs matching the candidate statistics & 8 & 4 & 8\\
\bottomrule
\end{tabularx}
\end{table}

The first two samples have disjoint factor orbits and were excluded from model training. Each pair shares its factors and has different output permutations with disjoint valid-answer sets. The eight earlier pairs also agree in the candidate statistics of Proposition~\ref{prop:schubert-candidate-data}. Component selection scanned all 96 attention heads and six multilayer perceptrons. The selected heads are $0,8$ in layer $1$ and $3,11$ in layer $3$, using zero-based indices. Random controls match their layers and the head pairing inherited from widening the model.

\begin{table}[H]
\centering\small
\caption{Confirmation results for Schubert interventions.}
\label{tab:schubert-interventions}
\begin{tabularx}{\linewidth}{@{}Yr@{}}
\toprule
Intervention & Mean score change\\
\midrule
Replace the four selected head outputs & 15.738\\
Twenty matched random head sets, averaged & 4.115\\
First-MLP activations outside the candidate set, through layer-1 keys and values & 11.658\\
Same replacement restricted to 32 previously selected hidden-unit pairs & 8.471\\
\quad Replace only the inverse-position contribution & 10.124\\
\quad Replace only the rank-matrix contribution & 0.211\\
\quad Replace only the descent-indicator contribution & 0.084\\
\quad Replace only the relative-root contribution & 0.470\\
Hold later MLP and attention outputs fixed during the 32-pair replacement & 0.435\\
\midrule
Ablation & Both answers correct\\
\midrule
No ablation & $248/249$\\
Zero the four selected heads & $215/249$\\
Zero all other attention heads & $80/249$\\
\bottomrule
\end{tabularx}
\end{table}

The selected-head effect exceeds the matched-control mean by $11.623$, with a descriptive $95\%$ bootstrap interval $[8.826,14.920]$. A hidden-unit pair consists of units $k,k+384$, inherited from widening the MLP; there are 384 such pairs. The input contributions are replaced before the activation function and subsequent layers, so their effects are not additive. The ablations leave MLPs and residual connections active. Together, the interventions show a contribution from inverse positions outside the candidate set and a dependence on later computation.

\subsubsection{Controlled ablation of model interpretation}
\label{app:interpretation-ablation}

\paragraph{Question and experimental setup.}
We asked whether access to mechanistic intervention evidence helps an agent
identify more influential model components and construct a mathematical rule
that generalizes better to unseen examples. We ran two GPT-6.1 Sol sessions,
each with a 60-minute budget. Both agents received the same tools, selector
outputs, and 320 component-selection pairs from the 29 factor orbits described
above. They also chose among the same 19 candidate sets of four attention heads
and used the available evidence to construct a mathematical rule for selecting
a valid move. The 19 sets are the distinct sets remaining after removing duplicates from the original selection and twenty matched random controls.

\paragraph{Ablated component.}
The only difference between the two conditions was access to information about
the effects of individual attention heads. The outputs-only agent received no
information about the selector's internal computation. In contrast, the
interpretation agent was also shown how the selector's prediction changed when
the output of each attention head for one example was replaced with its output
for the paired example.

\begin{table}[H]
\centering
\small
\caption{Controlled ablation of model-interpretation evidence.}
\label{tab:interpretation-ablation}
\begin{tabularx}{\linewidth}{@{}Ycc@{}}
\toprule
Result & Outputs only &
\shortstack{With head-intervention\\evidence} \\
\midrule
Selected head-set rank & 2 & 1 \\
Mean joint-replacement contrast & 10.46 & 16.18 \\
Inputs assigned a valid move & $423/498$ & $457/498$ \\
Pairs with both moves valid & $174/249$ & $214/249$ \\
Orbit-weighted pair success & $70.8\%$ & $80.3\%$ \\
\bottomrule
\end{tabularx}
\end{table}

\paragraph{Verification Note.} The accompanying script checks the head-set rankings and paired success counts using saved results. It can also rerun the two mathematical rules on the held-out examples, but it does not repeat the attention head swaps used to measure changes in the model's predictions.

\paragraph{Evaluation.}
We evaluated both conditions on the 249 confirmation pairs from 22 factor
orbits that were not used for component selection. In
Table~\ref{tab:interpretation-ablation}, the first two rows evaluate the head
set selected by each agent. These scores use the 248 pairs from 21 factor orbits on which the unmodified selector chooses valid moves for both inputs. Rank one indicates the strongest joint effect among the 19 candidate sets, and the mean joint-replacement contrast reports the size of that effect, measured by Equation~\eqref{eq:patch} and averaged equally over factor orbits. The remaining rows evaluate the agents' mathematical rules across the 498 individual inputs and the 249 pairs. The orbit-weighted success rate gives equal importance to each factor orbit rather than allowing larger orbits to dominate the result.
\\
The outputs-only agent chose the first candidate because the unmodified outputs did not distinguish attention head sets. The intervention-guided agent chose the set with the largest summed individual-head effects.
\\
Both mathematical rules used Schubert Newton polytope inequalities. The intervention-guided rule also used bounded descent cycling and the number of tight inequalities to choose among surviving moves. Both mathematical rules succeeded on the same 168 pairs. The
intervention-guided rule additionally succeeded on 46 pairs that the
outputs-only rule missed, whereas the outputs-only rule uniquely succeeded on
six pairs. Thus, the outputs-only condition was still able to construct a
useful rule, but access to the intervention results led to a stronger choice of model components and better performance on held-out examples. Each condition used one agent session.

\subsubsection{Ablation of hard-negative mining}
\label{app:hard-negative-ablation}

\paragraph{Question and setup.}
We examined what happens when hard-negative mining is omitted from model training in the Schubert investigation. In this condition, the tree was trained without actively selecting vanishing examples that passed the current necessary conditions. Unlike the controlled interpretation ablation in Appendix~\ref{app:interpretation-ablation}, the later research trajectories and human guidance were not matched exactly, so this experiment should be interpreted as a case-level ablation rather than a fully controlled comparison.

\paragraph{Results.}
Without hard-negative mining, the learned tree emphasized common descent statistics, and the agent identified six vanishing examples that survived descent cycling. Four of these examples could nevertheless be detected by retaining one flag dimension, while the remaining two could be detected by retaining two dimensions. The investigation ended with these finite counterexamples and an open multiplicity-one conjecture.

By contrast, the full investigation used hard-negative mining followed by further mathematical investigation and human redirection. It produced the recursive construction in Theorem~\ref{thm:schubert-projection}, showing that no fixed number of retained flag dimensions is sufficient. Thus, the ablation does not establish that hard-negative mining was solely responsible for the stronger result, but it provides evidence that mining helped redirect the investigation from isolated counterexamples toward a general structural obstruction.

In a separate matched comparison in $S_7$, ordinary sampling found six hard negatives per 2,048 coefficient evaluations, compared with 432 for uniform sampling among triples satisfying the symmetric Bruhat and contiguous-interval inversion inequalities. Two GPT-6.1 Sol extra-high agents received the same 4,096 training and 1,024 validation examples in $S_5$ and $S_6$, their 2,048 additional $S_7$ labels, and a depth-six decision tree fitted to their data, with a 60-minute budget each. Their rules certified $188/190$ and $189/190$ held-out zeros satisfying the inequalities in $S_7$ and $S_8$, respectively, and neither certified any of the 578 positive coefficients as zero. Mining thus supplied more hard negatives, while the resulting rules differed on one held-out zero.

\subsubsection{Rediscovered necessary conditions}
\label{app:schubert-classical}
\begin{proposition}[{\citealp{knutson2003_recurrence}}]
\label{prop:schubert-descent}
If $c_{u,v}^{w}>0$, then $\Des(w)\subseteq\Des(u)\cup\Des(v)$.
\end{proposition}
The earlier model experiments suggested this descent condition, which the system then used to select hard negatives for further training.

For $x\in S_n$, let $r_x(i,j)=\#\{a\leq i:x(a)\leq j\}$. The \emph{Rothe diagram} is $D(x)=\{(i,j):j<x(i),\ i<x^{-1}(j)\}$, and its \emph{essential set} consists of the southeast corners. The following condition is a consequence of Grassmannian projection and the Littlewood--Richardson rule \citep{fulton1997}.
\begin{proposition}[{\citealp{fulton1997}}]
\label{prop:schubert-rank}
If $(i,j)\in\operatorname{Ess}(w)$ and $r_u(i,j)=r_v(i,j)=r_w(i,j)$, then $c_{u,v}^{w}=0$.
\end{proposition}
A model trained on triples passing the earlier conditions prompted a search over rank-matrix features, yielding the equal-rank condition above. It excluded 133 of 45,000 hard negatives, with no violations among 59,979 nonzero coefficients (Table~\ref{tab:schubert-mining}).

\subsubsection{Vanishing under proper projections}
\label{app:schubert-projections}

For $E\subseteq\{1,\ldots,n-1\}$, let $\operatorname{Fl}(E;\C^n)$ retain the subspaces with dimensions in $E$. Write $p_E(x)$ for the minimal representative obtained by sorting the entries of $x$ within the position blocks determined by $E$. The generic projected intersection is nonempty exactly when
\[
\sigma_{p_E(u)}\sigma_{p_E(v)}\sigma_{p_E(w_0w)}\ne0.
\]
These are pullbacks of partial-flag Schubert classes. If $D=\Des(u)\cup\Des(v)$ and $\Des(w)\subseteq D$, the product at $E=D$ is nonzero exactly when $c_{u,v}^{w}>0$: the first two Schubert conditions pull back from this partial flag variety, and the projection of the third is indexed by $p_D(w_0w)$.

\begin{theorem}[LLM-generated]
\label{thm:schubert-projection}
For every $n\geq5$, there exist $u,v,w\in S_n$ with $\ell(w)=\ell(u)+\ell(v)$, $c_{u,v}^{w}=0$, and
\[
\Des(w)\subseteq D=\Des(u)\cup\Des(v),\qquad |D|=n-3,
\]
such that every projection retaining a proper subset $E\subsetneq D$ has a nonempty generic intersection.
\end{theorem}

Starting with $(u,v,w)=(215346,135264,256134)$, let $f$ be the factor beginning with $1$ and $g$ the other factor. To form $f'$ and $w'$, prepend $2$ to $f$ and $w$ and increase each old value at least $2$ by one. To form $g'$, prepend $1$ to $g$ and increase every old value by one. Preserve the labels of the two factors at each step. This recursion supplies the examples for every $n\geq6$; $(14235,24153,45123)$ supplies $n=5$.

\paragraph{The projected intersection matrix}
For three generic upper unitriangular matrices $\rho_1,\rho_2,\rho_3$, form the matrix $F_E$ whose columns are
\[
\rho_i e_{ab}\rho_i^{-1},\qquad a<b,\ x_i(a)>x_i(b),\qquad
(x_1,x_2,x_3)=(p_E(u),p_E(v),p_E(w_0w)).
\]
Rows are indexed by upper triangular positions $(s,t)$, and the entry in column $(i,a,b)$ is $\rho_i[s,a]\rho_i^{-1}[b,t]$. Purbhoo's criterion identifies nonemptiness of the generic projected intersection with full column rank of $F_E$ \citep{purbhoo2003_root_game}. All subsequent ranks and kernels are taken over the rational function field in the matrix entries.

Set $q=(p_D(w_0w))^{-1}(n)$, the position of the maximum after sorting within the parabolic blocks. For the seed in $S_6$, $p_D(w_0w)=(5,1,2,4,6,3)$ and $q=5$.
\begin{lemma}[LLM-generated]
\label{lem:schubert-propagation}
Suppose $(u,v,w)\in S_n^3$ has $\ell(w)=\ell(u)+\ell(v)$, has empty generic intersection on $D$, and has nonempty generic intersection on every proper subset of $D$, with $\Des(w)\subseteq D$ and $|D|=k$. Assume that $1\in D$, $w(1)=2$, and exactly one of $u(1),v(1)$ equals 1. Suppose $F_{D}$ has column nullity one, with kernel vector $\kappa$, and, for the index $q$ defined above, at least one of
\begin{equation}
\Psi_{i,a}=\sum_{b>a}\kappa_{(i,a,b)}
                 (\rho_i^{-1}\rho_3)[b,q],\qquad i\in\{1,2\},
\label{eq:schubert-functional}
\end{equation}
is a nonzero rational function. Coordinates absent from the matrix are zero. Every $t$-fold application of this recursion then has empty generic intersection on its $k+t$ descent dimensions in $S_{n+t}$ and nonempty generic intersection on every proper subset.
\end{lemma}

\begin{proof}
Write $f$ for the factor beginning with 1 and $g$ for the other factor. The construction sends $(f,g,w)$ to $(f',g',w')$, with the two factor labels preserved. It increases the union of descents to $D'=\{1\}\cup(D+1)$. The first two positions are singleton blocks for $D'$, and the hypotheses on the first entries persist.

For any $E\subseteq D$, let $E'=\{1\}\cup(E+1)$ and extend the generic flags as $\rho'_i=\left[\begin{smallmatrix}1&r_i\\0&\rho_i\end{smallmatrix}\right]$, where the $r_i$ have independent entries. Sorting within blocks commutes with this extension. Order rows and columns with first position 1 first. The matrix takes the form
\begin{equation}
F_{E'}=\begin{pmatrix}X&Y\\0&F_E\end{pmatrix}.
\label{eq:schubert-block}
\end{equation}
There are $n$ rows and $n$ columns involving the new first position. The square block $X$ represents the projective-space intersection with codimensions $1,0,n-1$, whose degree is one. Therefore $X$ is generically invertible. This is the degree-one base intersection in parabolic factorization \citep{richmond2008_multiplicative}. Equation~\eqref{eq:schubert-block} preserves full column rank and column nullity. In particular it preserves vanishing for the full flag problem, and it preserves feasibility for every proper projection containing the newly retained dimension.

It remains to check $M=D+1$, which omits that new dimension. Every other proper projection omitting it is a projection of $M$. Passing from $D'$ to $M$ sorts only positions 1 and 2. In $f'$ these entries are 2 and 1; in the third factor they are $n$ and $n-1$. Their values are consecutive, so sorting removes the inversion $(1,2)$ and changes no other inversion. The factor $g'$ is already increasing at those positions. Thus
\begin{equation}
F_M=F_{D'}\text{ with columns }(f,1,2)\text{ and }(3,1,2)\text{ deleted}.
\label{eq:schubert-deletions}
\end{equation}

Put $F=F_{D}$. Its kernel lifts to $\kappa'=(-X^{-1}Y\kappa,\kappa)$. On row $(1,2)$, the two columns to be deleted both have entry one, all other columns involving the first position have entry zero, and $Y\kappa$ has entry zero. Hence their kernel coordinates $\alpha_f$ and $\alpha_3$ are opposite. Since the kernel is one-dimensional, Equation~\eqref{eq:schubert-deletions} has full column rank exactly when $\alpha_f\ne0$.

The position of the maximum in the new third minimal representative is $q'=q+1$. Its inversions involving the first position include $(1,j)$ for all $j\geq2$ except $j=q'$. If $\alpha_f=\alpha_3=0$, the remaining coordinates involving the first position must solve the equations on rows $(1,j)$, $j\geq3$, using the rows of $\rho_3'^{-1}$ indexed by $j\geq3$, $j\ne q'$. Their annihilator is the $q'$-th column of the corresponding lower-right block of $\rho'_3$. It follows that the equations are solvable exactly when
\begin{equation}
\Phi=\sum_{i=1}^{2}\sum_{(a,b)\in\operatorname{Inv}(x_i)}
\kappa_{(i,a,b)}\,\rho'_i[1,a+1]
(\rho_i^{-1}\rho_3)[b,q]=0.
\label{eq:schubert-extension-functional}
\end{equation}
There is no contribution from the third factor: such a term would require an inversion ending at the position of its maximum. By Equation~\eqref{eq:schubert-functional}, the coefficient of the fresh variable $\rho'_i[1,a+1]$ is $\Psi_{i,a}$. At least one is nonzero, so $\Phi$ is not identically zero and $F_M$ is generically of full column rank.

Finally, the nonzero rational function persists at every extension. Normalize each lifted kernel so that its restriction to old columns equals the previous kernel. After $t$ extensions a seed column $(i,a,b)$ has become $(i,a+t,b+t)$, and its coordinate is unchanged. The corresponding lower-right blocks of $\rho_i^{-1}\rho_3$ are also unchanged, while $q_t=q+t$. Every newly added column has first position at most $t$; therefore none contributes to the sum with first position $a+t>t$. Consequently $\Psi^{(t)}_{i,a+t}=\Psi^{(0)}_{i,a}$. The same fresh-variable argument proves feasibility of $M$ at every stage. Together with the proper projections containing the new dimension, this proves the assertion by induction.
\end{proof}

\paragraph{Seed calculations}
For the $S_6$ seed, divided differences give $c_{215346,135264}^{256134}=0$. Table~\ref{tab:schubert-seeds} supplies a nonzero term for every nonempty proper projection. In each row, $c_{p_E(u),p_E(v)}^z=1$ and $z\leq w_0p_E(w_0w)$, so the projected triple product is nonzero. The same calculation applies to the $S_5$ seed, whose ordinary coefficient is also zero.

\begin{table}[H]
\centering\small
\caption{Nonzero terms in the proper projections of the two Schubert seeds.}
\label{tab:schubert-seeds}
\begin{tabular}{@{}clcc@{}}
\toprule
$n$ & Retained dimensions $E$ & $z$ & $w_0p_E(w_0w)$\\
\midrule
6 & $\{1\}$ & $213456$ & $265431$\\
6 & $\{3\}$ & $146235$ & $652431$\\
6 & $\{5\}$ & $123564$ & $653214$\\
6 & $\{1,3\}$ & $246135$ & $265431$\\
6 & $\{1,5\}$ & $213564$ & $265314$\\
6 & $\{3,5\}$ & $236154$ & $652314$\\
5 & $\{2\}$ & $35124$ & $54321$\\
5 & $\{4\}$ & $12453$ & $54213$\\
\bottomrule
\end{tabular}
\end{table}

The rank and kernel conditions for the $S_6$ seed can be checked at the following three upper unitriangular matrices. Listing their strictly upper triangular entries in lexicographic order of $(a,b)$ gives
\[
\begin{aligned}
(\rho_1[a,b])_{a<b}&=(32,25,46,17,25,81,60,45,33,2,24,62,15,65,28),\\
(\rho_2[a,b])_{a<b}&=(53,56,78,26,29,97,53,91,54,35,51,11,6,94,54),\\
(\rho_3[a,b])_{a<b}&=(95,10,36,64,23,65,34,7,11,16,25,47,37,98,47).
\end{aligned}
\]
Order rows of $F_D$ by $(a,b)$ and columns by $(i,a,b)$, both lexicographically. Deleting rows $(2,3),(3,5),(4,5)$ and column $(3,5,6)$ gives a $12\times12$ minor of determinant $-349758000$. Vanishing bounds the generic rank of this $15\times13$ matrix by twelve, so its generic column nullity is one. Normalize the kernel by $\kappa_{(3,5,6)}=1$. At this specialization, $q=5$ and $\Psi_{1,3}=-191797/254\ne0$. Cramer's rule expresses this normalized kernel near the specialization, proving that $\Psi_{1,3}$ is not identically zero. Lemma~\ref{lem:schubert-propagation} now proves Theorem~\ref{thm:schubert-projection} for $n\geq6$, and the separate seed proves it for $n=5$.

\subsubsection{Equivariant positivity and nonvanishing}
\label{app:schubert-equivariant}

The \emph{double and triple Schubert structure constants} are defined by
\begin{align*}
\mathfrak S_u(x;t)\mathfrak S_v(x;t)&=\sum_w c_{u,v}^{w}(t)\mathfrak S_w(x;t),\\
\mathfrak S_u(x;y)\mathfrak S_v(x;t)&=\sum_w c_{u,v}^{w}(t;y)\mathfrak S_w(x;t).
\end{align*}
The first coefficients are the structure constants of $H_T^*(\operatorname{Fl}_n;\Z)$ when $u,v,w\in S_n$. The second multiplication uses two independent parameter sets and need not be symmetric in $u,v$. Set $\alpha_i=t_{i+1}-t_i$ and $\Phi^+=\{\alpha_a+\cdots+\alpha_b:1\leq a\leq b<n\}$. Graham positivity gives $c_{u,v}^{w}(t)\in\Z_{\geq0}[\alpha_1,\ldots,\alpha_{n-1}]$ \citep{graham1999_positivity}.

\begin{theorem}[LLM-generated]
\label{thm:schubert-positivity}
For $u,v,w\in S_n$ with $s_i u<u\leq w$ in Bruhat order, the difference
\[
c_{u,v}^{w}(t)-\alpha_i c_{s_i u,v}^{w}(t)
\]
is a nonnegative integer combination of products of distinct roots in $\Phi^+\setminus\{\alpha_i\}$.
\end{theorem}

\begin{corollary}[LLM-generated]
\label{cor:schubert-support}
For $u,v,w\in S_\infty$,
\begin{align}
c_{u,v}^{w}(t)\ne0
&\ \Longleftrightarrow\ u\leq w\ \text{and}\ c_{u,v}^{w}(t;y)\ne0,
\label{eq:schubert-support-one}\\
&\ \Longleftrightarrow\ c_{u,v}^{w}(t;y)\ne0\ \text{and}\ c_{v,u}^{w}(t;y)\ne0.
\label{eq:schubert-support-both}
\end{align}
Consequently Conjectures A and B of \citet{chen2026_support} hold.
\end{corollary}

\begin{proof}
By Theorem~2.2 of \citet{chen2026_support}, $c_{u,v}^{w}(t;y)\ne0$ if and only if there is a length-additive factorization $u=az$ with $c_{z,v}^{w}>0$. If $u\leq w$, every element of the associated chain in left weak order from $z$ to $u$ lies below $w$. The coefficient $c_{z,v}^{w}(t)$ is the positive integer $c_{z,v}^{w}$, since its degree is zero. Iterating Theorem~\ref{thm:schubert-positivity} gives a nonzero monomial lower bound for $c_{u,v}^{w}(t)$. Conversely, specialization $y=t$ and the necessary Bruhat bound prove the reverse implication in Equation~\eqref{eq:schubert-support-one}.

The same description of triple nonvanishing gives $c_{v,u}^{w}(t;y)\ne0\Rightarrow u\leq w$. Equation~\eqref{eq:schubert-support-one} therefore proves the reverse implication in Equation~\eqref{eq:schubert-support-both}; specialization and symmetry of double multiplication prove the forward implication. This is Conjecture B. By Theorem~1.1 of \citet{chen2026_support}, it is equivalent to Conjecture A, which states that $c_{u,v}^{w}(t)\ne0$ exactly when there are length-additive factorizations $u=u_1u_2$ and $v=v_1v_2$ with $c_{u,v_2}^{w}>0$ and $c_{u_2,v}^{w}>0$.
\end{proof}

\paragraph{Geometric setup for the proof}
Let $G=\mathrm{GL}_n(\C)$, let $B,B^-$ be opposite \emph{Borel subgroups}, and let $T=B\cap B^-$. Choose the identification of the \emph{character lattice} with the parameters $t$ so that the weights of $U=R_u(B)$ are $\Phi^+$. For upper triangular matrices, this means that the $a,b$ entry, $a<b$, has weight $t_b-t_a$. Put $N=\dim G/B$ and
\[
C_w=BwB/B,\qquad X_w=\overline{C_w},\qquad
X^u=\overline{B^-uB/B},\qquad \sigma_u=[X^u]_T.
\]
We use \emph{equivariant Chow classes}, identified here with equivariant cohomology classes. The coefficient can be extracted by
\begin{equation}
c_{u,v}^{w}(t)=\int_{X_w}\sigma_u\sigma_v.
\label{eq:schubert-integral}
\end{equation}
Indeed, $\int_{X_w}\sigma_y=\delta_{y,w}$. If $\ell(y)<\ell(w)$, the integral has negative degree. Otherwise $X^y\cap X_w$ is empty unless $y=w$, in which case the intersection is the transverse fixed point.

\paragraph{Effective cycles and multidegrees}

Root coordinates identify $U$ with an affine $T$-representation, with one coordinate of each vector weight $\beta\in\Phi^+$. For a closed $T$-invariant subvariety $Y$ of such a representation $V$, write $\operatorname{mdeg}_T^V(Y)=[Y]_T\in A_T^*(V)=A_T^*(\mathrm{pt})$. We omit the superscript when the ambient representation is specified.

\begin{lemma}[{\citealp[Theorem~D]{knutsonmiller2005}}]
\label{lem:schubert-multidegree}
An effective $T$-invariant cycle in a representation with positive-root weights has multidegree equal to a nonnegative integer combination of products of distinct coordinate weights.
\end{lemma}
This classical lemma supplies the multidegree positivity used in the proof of Theorem~\ref{thm:schubert-positivity}. In particular, every nonempty invariant subvariety has nonzero multidegree: a product of positive roots is nonzero and has nonnegative simple-root coefficients.

\paragraph{Intersection over a minimal parabolic subgroup}

Fix $i$, let $P_i$ be the \emph{minimal parabolic} containing $B$, and put $K_i=R_u(P_i)$ and $d=N-1$. Its coordinate weights are precisely $\Phi^+\setminus\{\alpha_i\}$. A representative $n_i$ of $s_i$ normalizes $K_i$.

Suppose $s_i z>z$. The $P_i$-orbit $\Omega_z=C_z\sqcup C_{s_i z}$ maps to $P_i/B\simeq\mathbb P^1$. The stabilizer of $zB$ in $P_i$ lies in $B$ and contains $T U_{\alpha_i}$, so $K_i$ acts transitively on each fiber, of dimension $\ell(z)$. Label the point whose fiber is $C_z$ by $\infty$ and the other fixed point by $0$. Then $C_{s_i z}$ lies over $\mathbb A^1=\mathbb P^1\setminus\{\infty\}$, with fiber $n_iC_z$ at $0$, and $n_i$ interchanges $0$ and $\infty$.

For any $h$, the image of $X^h\cap\Omega_z$ is
\begin{equation}
\begin{cases}
\mathbb P^1,&h\leq z,\\
\{0\},&h\leq s_i z,\ h\not\leq z,\\
\varnothing,&h\not\leq s_i z.
\end{cases}
\label{eq:schubert-orbit-image}
\end{equation}
Choose a cocharacter that vanishes on $\alpha_i$ and is positive on all other positive roots. Its limit contracts each fiber to the Levi fixed component. The variety $X^h$ is closed and $T$-stable. Its intersection with this component is $B_i^-$-stable and has fixed points given by Bruhat order, which gives the three stated possibilities. The zero section supplies every point of the indicated image.

We use the standard Richardson-variety fact that $X^h\cap C_y$, when nonempty, is irreducible of dimension $\ell(y)-\ell(h)$. If $h\leq z$, the affine variety $X^h\cap C_{s_i z}$ surjects onto $\mathbb A^1$. Every fiber has dimension $\ell(z)-\ell(h)$: it is cut out in an integral affine variety by one nonconstant regular function minus a scalar. The fiber at $\infty$, namely $X^h\cap C_z$, has the same dimension. If $h\leq s_i z$ but $h\not\leq z$, the whole patch lies over $0$ and has dimension $\ell(z)+1-\ell(h)$.

Consider the closed incidence scheme
\begin{equation}
\mathcal Z=\{(g,F)\in K_i\times X_w:
 F\in n_iX^u,\quad g^{-1}F\in X^v\}.
\label{eq:schubert-incidence}
\end{equation}
For its part $\mathcal Z_y$ over $C_y$, put $D_y=d+\ell(y)-\ell(u)-\ell(v)$. The nonempty strata have dimensions
\begin{equation}
\begin{array}{c|l|c}
 &\text{conditions}&\dim\mathcal Z_y\\\hline
s_i y>y&u\leq y,\ v\leq y&D_y\\
s_i y>y&u\not\leq y,\ u\leq s_i y,\ v\leq y&D_y+1\\
s_i y<y&u\leq s_i y,\ v\leq y&D_y.
\end{array}
\label{eq:schubert-dimensions}
\end{equation}
All other strata are empty. Here is the parameter count. If $s_i y>y$, the cell is one $K_i$-orbit. Translation by $n_i$ identifies $n_iX^u\cap C_y$ with the fiber at $0$ of $X^u\cap C_{s_i y}$. Its dimension is respectively $\ell(y)-\ell(u)$ or $\ell(y)+1-\ell(u)$ in the first two rows. For each such $F$, the choices of $g$ have dimension $d-\ell(v)$.

If $s_i y<y$, put $z=s_i y$. When $u\not\leq z$, the image of $X^u$ in Equation~\eqref{eq:schubert-orbit-image} is empty or $\{0\}$, whereas $n_iC_y$ avoids that fiber; hence the stratum is empty. When $u\leq z$, every fiber of $n_iX^u\cap C_y$ has dimension $\ell(y)-\ell(u)-1$. If $v\leq z$, the choices of $g$ have dimension $d-\ell(v)$ and the base point varies in $\mathbb A^1$. If $v\not\leq z$ but $v\leq y$, only base point $0$ is allowed and the choices of $g$ have dimension $d+1-\ell(v)$. Both give the last row.

Under the hypothesis $u\leq w$, a stratum in the second row has $y<w$. Thus $\ell(y)+1\leq\ell(w)$, and every stratum has dimension at most $d+\ell(w)-\ell(u)-\ell(v)$. This is the expected dimension in Equation~\eqref{eq:schubert-incidence}. The intersection dimension inequality supplies the reverse bound at every nonempty component. Consequently the three subvarieties defining this incidence scheme in the smooth variety $K_i\times G/B$ meet properly.

\begin{proof}[Proof of Theorem~\ref{thm:schubert-positivity}]
The proper intersection above defines an effective intersection cycle. The Schubert varieties and their indicated inverse images are Cohen--Macaulay; the expected codimension ensures positive intersection multiplicities at the generic points of the components. Projection $\pi:K_i\times X_w\to K_i$ is proper. The action map and the projection $K_i\times G/B\to G/B$ induce the same pullback in equivariant Chow theory. In type A an equivariant homotopy is given by $g\mapsto 1+t(g-1)$ in root coordinates, followed by the action on the flag variety. Also the left Demazure identity gives
\[
[n_iX^u]_T=s_i\sigma_u
=\sigma_u-\alpha_i\sigma_{s_i u}\qquad(s_i u<u).
\]
Therefore
\[
\pi_*[\mathcal Z]_{\mathrm{int}}
=\int_{X_w}(s_i\sigma_u)\sigma_v
=c_{u,v}^{w}(t)-\alpha_i c_{s_i u,v}^{w}(t).
\]
Proper pushforward preserves effective cycles: components with positive-dimensional generic fibers contribute zero, and the others contribute their positive generic degrees. Apply Lemma~\ref{lem:schubert-multidegree} to the representation $K_i$.
\end{proof}

\subsubsection{A limitation of the statistics at candidate indices}
\begin{proposition}[LLM-generated]
\label{prop:schubert-candidate-data}
For every $N\geq10$, there exist $u,v,w,w'\in S_N$ with a common supplied left descent and nonempty disjoint answer sets, for which the following data agree at every $j\in\Des_L(v)$:
\[
 \begin{gathered}
 z^{-1}(j),\ z^{-1}(j+1),\qquad r_z(a,j),\ r_z(a,j+1)\quad(1\leq a\leq N),\\
 (xy^{-1})(e_{j+1}-e_j)\quad\text{for ordered distinct }x,y\in\{u,v,z\},
 \end{gathered}
\]
where the comparison is between $z=w$ and $z=w'$. No deterministic rule depending only on these data and the common factors can choose a valid answer for both inputs.
\end{proposition}
\begin{proof}
Take
\[
\begin{aligned}
 u&=(1,3,8,2,6,4,7,5,9,10),&v&=(1,6,2,3,7,5,8,4,9,10),\\
 w&=(1,10,5,2,6,4,8,3,7,9),&w'&=(2,10,5,1,6,4,7,3,8,9).
\end{aligned}
\]
Their lengths are $9,8,16,16$, respectively, and $\Des_L(v)=\{4,5\}$. Schubert multiplication gives
\[
\begin{array}{c|ccc}
 z&c_{s_5u,v}^{z}&c_{u,s_4v}^{z}&c_{u,s_5v}^{z}\\\hline
 w&1&0&2\\
 w'&1&1&0
\end{array}
\]
so $\mathcal V(u,v,w)=\{5\}$ and $\mathcal V(u,v,w')=\{4\}$. The displayed statistics agree by direct substitution. The same inputs to a deterministic rule give the same output, which cannot belong to both sets. Adjoining fixed points preserves the coefficients and all stated equalities, proving the assertion for every $N\geq10$.
\end{proof}

The model distinguishes inputs with identical candidate statistics by using other positions. Table~\ref{tab:schubert-equal-data} records the predictions when first-MLP activations outside the candidate set are transferred through layer-1 keys and values.

\begin{table}[H]
\centering\small
\caption{Interventions on inputs with identical candidate statistics.}
\label{tab:schubert-equal-data}
\begin{tabularx}{\linewidth}{@{}Yl@{}}
\toprule
Input or intervention & Model answer\\
\midrule
Unmodified $w$, $w'$ &5, 4\\
Transfer from $w$ to $w'$ &$4\to5$\\
Transfer from $w'$ to $w$ &$5\to5$\\
All eight pairs with identical candidate statistics &Donor-valid answer in 12/16 directions\\
\bottomrule
\end{tabularx}
\end{table}
\end{pipelineoutput}

\FloatBarrier
\Needspace{12\baselineskip}
\subsection{Kronecker coefficients}
\label{app:kronecker}

\subsubsection{Mathematical setting}
Partitions $\lambda\vdash n$ index the irreducible complex representations $V^\lambda$ of $S_n$. The Kronecker coefficient is the tensor-product multiplicity
\[
 V^\lambda\otimes V^\mu\cong\bigoplus_{\nu\vdash n}(V^\nu)^{\oplus g_{\lambda,\mu,\nu}}.
\]
We follow the character and symmetric-function conventions of \citet{sagan2001}. If $\rho=(1^{m_1}2^{m_2}\cdots)$ and $z_\rho=\prod_i i^{m_i}m_i!$, character orthogonality gives the formula
\[
 g_{\lambda,\mu,\nu}=\sum_{\rho\vdash n}\frac{\chi^\lambda(\rho)\chi^\mu(\rho)\chi^\nu(\rho)}{z_\rho}.
\]
The implementation evaluates characters by the \emph{Murnaghan--Nakayama rule} and checks them against the \emph{hook-length formula} and orthogonality. Figure~\ref{fig:kronecker} shows $V^{(2,1)}\otimes V^{(2,1)}\cong V^{(3)}\oplus V^{(2,1)}\oplus V^{(1,1,1)}$.

\begin{pipelineoutput}
\begin{figure}[H]
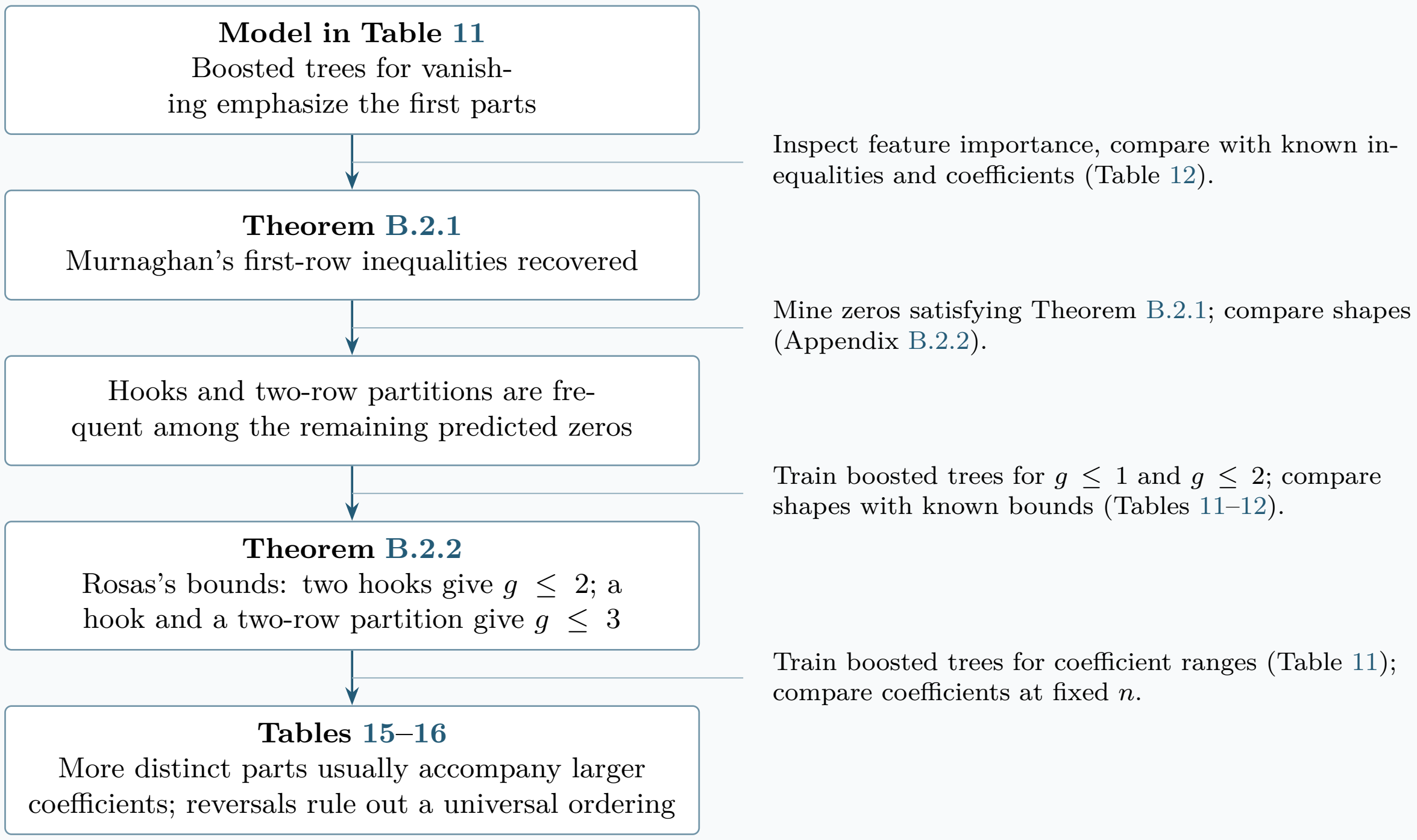

\centering
\begin{discoverychart}
\discoveryfirst{k1}{\textbf{Model in Table~\ref{tab:kronecker-models}}\\Boosted trees for vanishing emphasize the first parts}
\discoverynext{k2}{k1}{\textbf{Theorem~\ref{thm:kronecker-murnaghan}}\\Murnaghan's first-row inequalities recovered}{Inspect feature importance, compare with known inequalities and coefficients (Table~\ref{tab:kronecker-exact}).}
\discoverynext{k3}{k2}{Hooks and two-row partitions are frequent among the remaining predicted zeros}{Mine zeros satisfying Theorem~\ref{thm:kronecker-murnaghan}; compare shapes (Appendix~\ref{app:kronecker-models}).}
\discoverynext{k4}{k3}{\textbf{Theorem~\ref{thm:kronecker-rosas}}\\Rosas's bounds: two hooks give $g\leq2$; a hook and a two-row partition give $g\leq3$}{Train boosted trees for $g\leq1$ and $g\leq2$; compare shapes with known bounds (Tables~\ref{tab:kronecker-models}--\ref{tab:kronecker-exact}).}
\discoverynext{k5}{k4}{\textbf{Tables~\ref{tab:kronecker-comparison}--\ref{tab:kronecker-correlation}}\\More distinct parts usually accompany larger coefficients; reversals rule out a universal ordering}{Train boosted trees for coefficient ranges (Table~\ref{tab:kronecker-models}); compare coefficients at fixed $n$.}
\end{discoverychart}
\caption{Kronecker rediscoveries and the subsequent empirical comparison.}
\label{fig:app-kronecker-flow}
\end{figure}

\subsubsection{Models and hard-negative experiments}
\label{app:kronecker-models}
Each input is an unordered triple of partitions, sorted lexicographically and encoded by its parts padded with zeros, together with $n$. We predict vanishing $(g=0)$, the thresholds $g \leq 1$ and $g \leq 2$, and a seven-class range of $g$: $g=0,\ g=1,\ g=2,\ \dots,\ g=5,\ g\geq6$.

Training uses all 1,095 such triples for $3\leq n\leq7$. Evaluation uses 6,984 triples at $n=8,9$ and 44,100 at $n=10,11$. Padding is extended before training a model evaluated at larger $n$. The retained gradient-boosting implementation uses 200 trees of maximum depth three. Table~\ref{tab:kronecker-models} reports classification accuracy.

\begin{table}[H]
\centering\small
\caption{Kronecker classification accuracy.}
\label{tab:kronecker-models}
\begin{tabular}{@{}lcrr@{}}
\toprule
Target & Evaluation sizes & Logistic regression & Boosted trees\\
\midrule
$g=0$ & $8,9$ & $77.51\%$ & $89.09\%$\\
$g\leq1$ & $8,9$ & $76.92\%$ & $87.41\%$\\
$g\leq1$ & $10,11$ & $72.11\%$ & $79.27\%$\\
$g\leq2$ & $8,9$ & $80.70\%$ & $87.41\%$\\
$g\leq2$ & $10,11$ & $69.84\%$ & $72.22\%$\\
$g\in\{0,1,\dots,5,{\geq}6\}$ & $10,11$ & $30.94\%$ & $46.46\%$\\
\bottomrule
\end{tabular}
\end{table}

Hard negatives for vanishing are zeros satisfying Murnaghan's inequalities in Theorem~\ref{thm:kronecker-murnaghan}. The first search also excluded triples containing $(n)$ or $(1^n)$, whose coefficients follow from character orthogonality. Among 5,600 remaining test triples at $n=8,9$, the classifier correctly identified 1,145 zeros that passed those inequalities. Their first and second parts suggested testing the analogous inequalities after removing two rows. Counterexamples rejected that extension; Proposition~\ref{prop:kronecker-truncation} gives a small example.

The same strategy applies to Rosas's bounds in Theorem~\ref{thm:kronecker-rosas}: both bounds hold on every sampled triple, and classifier predictions for the magnitude tasks respect them almost everywhere. Table~\ref{tab:kronecker-exact} summarizes all four checks.

\begin{table}[H]
\centering\small
\caption{Kronecker tests of necessary conditions and coefficient bounds.}
\label{tab:kronecker-exact}
\begin{tabularx}{\linewidth}{@{}Y>{\raggedright\arraybackslash}p{.25\linewidth}r@{}}
\toprule
Condition or prediction & Population & Violations\\
\midrule
Murnaghan's inequalities on coefficients & $3\leq n\leq9$; 1,479 exclusions & $0$\\
Triangle inequalities after deleting two rows & Same sizes; 4,014 exclusions & $1{,}344$ positive coefficients\\
Two hooks imply $g\leq2$ & 2,308 triples & $0$\\
A hook and a two-row partition imply $g\leq3$ & 2,262 triples & $0$\\
Classifier prediction respects the two-hook bound & 5,236 triples at $n=10,11$ & $0$\\
Classifier prediction respects the hook/two-row bound & 5,232 triples at $n=10,11$ & $15$\\
\bottomrule
\end{tabularx}
\end{table}

A further check asked whether classifier predictions for the magnitude task respect the semigroup property of Theorem~\ref{thm:kronecker-semigroup}. The theorem was confirmed on 35 sampled pairs with positive coefficients. Predictions for the 1,614 sampled sums satisfied this inequality more often than a size-matched baseline, with a small improvement ($78.1\%$ vs.\ $74.2\%$) (Table~\ref{tab:kronecker-semigroup}). Much of the agreement reflects the baseline effect that predictions increase with $n$, and a sum always has a larger $n$ than either summand.

\begin{table}[H]
\centering\small
\caption{Semigroup-inequality agreement for magnitude predictions.}
\label{tab:kronecker-semigroup}
\begin{tabular}{@{}lr@{}}
\toprule
 & Agreement\\
\midrule
Classifier predictions & 78.1\%\\
Size-matched baseline & 74.2\%\\
\bottomrule
\end{tabular}
\end{table}

Returning to Rosas's bounds directly, after removing the triples covered by either bound, the dataset leaves 348 training triples and 35,730 evaluation triples at $n=10,11$. Table~\ref{tab:kronecker-residual} reports retraining on this population. Agreement with a coefficient bound in Table~\ref{tab:kronecker-exact} measures a weaker requirement than predicting the coefficient itself.

\begin{table}[H]
\centering\small
\caption{Kronecker prediction outside the two special-shape bounds.}
\label{tab:kronecker-residual}
\begin{tabular}{@{}lr@{}}
\toprule
Predictor for $g\leq2$ after removing Rosas's cases & Accuracy\\
\midrule
Retrained boosted trees &59.72\%\\
Constant majority prediction &53.23\%\\
\bottomrule
\end{tabular}
\end{table}

\subsubsection{Rediscovered coefficient bounds}
\label{app:kronecker-results}
\begin{theorem}[{\citealp[Equation~(22)]{mishna2021}}]
\label{thm:kronecker-murnaghan}
If $g_{\lambda,\mu,\nu}>0$, then
\[
 n-\lambda_1\leq(n-\mu_1)+(n-\nu_1),
\]
and the two inequalities obtained by permuting $\lambda,\mu,\nu$ also hold.
\end{theorem}
Feature importance in the boosted trees emphasized the first parts, prompting comparison with Murnaghan's inequalities. The system then mined zeros satisfying these inequalities to investigate the remaining predictions.

\begin{theorem}[{\citealp[Corollaries~15 and~17]{rosas2001}}]
\label{thm:kronecker-rosas}
For $\lambda,\mu,\nu\vdash n$, the following bounds hold, with the third partition arbitrary:
\[
\begin{aligned}
 g_{\lambda,\mu,\nu}&\leq2&&\text{if two partitions are hooks},\\
 g_{\lambda,\mu,\nu}&\leq3&&\text{if two partitions are a hook and a two-row partition}.
\end{aligned}
\]
A hook is a partition $(n-r,1^r)$.
\end{theorem}
Hard-negative mining found an excess of hooks and two-row partitions among correctly predicted zeros. Subsequent coefficient-threshold experiments led the agent to Rosas's bounds, which it checked by computation and against the model predictions (Table~\ref{tab:kronecker-exact}).

\begin{theorem}[{\citealp[Proposition~2.4]{bessenrodt2017}}]
\label{thm:kronecker-semigroup}
If $g_{\lambda,\mu,\nu}>0$ and $g_{\alpha,\beta,\gamma}>0$, then
\[
 g_{\lambda+\alpha,\mu+\beta,\nu+\gamma}\geq\max(g_{\lambda,\mu,\nu},g_{\alpha,\beta,\gamma}).
\]
\end{theorem}
The system used this known inequality to test the magnitude predictions on sums of partition triples (Table~\ref{tab:kronecker-semigroup}).

\begin{proposition}[LLM-generated]
\label{prop:kronecker-truncation}
The analogous triangle condition on $n-\lambda_1-\lambda_2$, $n-\mu_1-\mu_2$, and $n-\nu_1-\nu_2$ is false.
\end{proposition}
\begin{proof}
For $\lambda=(1,1,1)$ and $\mu=\nu=(2,1)$, these three quantities are $1,0,0$. The tensor-product decomposition above gives $g_{\lambda,\mu,\nu}=1$.
\end{proof}

\subsubsection{Empirical comparison by the number of distinct parts}
For each triple, we compare the coefficient with
\[
 \#\{\lambda_i:\lambda_i>0\}+\#\{\mu_i:\mu_i>0\}+\#\{\nu_i:\nu_i>0\}.
\]
Pairs of triples are compared within the same $n$, omitting equal values of this statistic. A reversal means that the triple with the larger statistic has the smaller coefficient. The $n=10,11$ experiment sampled 30,223 comparable pairs; the later experiment sampled 6,000 triples at $n=15$.

\begin{table}[H]
\centering\small
\caption{Reversals in the comparison by distinct parts.}
\label{tab:kronecker-comparison}
\begin{tabular}{@{}rrrr@{}}
\toprule
Difference & \shortstack[r]{Pairs\\$n=10,11$} & \shortstack[r]{Reversals\\$n=10,11$} & \shortstack[r]{Reversals\\$n=15$}\\
\midrule
$1$ & 16,170 & $18.86\%$ & $25.48\%$\\
$2$ & 9,329 & $7.49\%$ & $12.04\%$\\
$3$ & 3,599 & $2.53\%$ & $5.47\%$\\
$\geq4$ & 1,125 & $0.44\%$ & $1.41\%$\\
\bottomrule
\end{tabular}
\end{table}

The comparisons in Table~\ref{tab:kronecker-comparison} concern individual triples. At a gap of one, reversals occur in roughly one in five comparisons at $n=10,11$ and one in four at $n=15$; at a gap of at least four, the respective rates are $0.44\%$ and $1.41\%$. These reversals nonetheless preclude an unconditional ordering by the displayed statistic.

\begin{table}[H]
\centering\small
\caption{Association with Kronecker coefficients at $n=10,11$.}
\label{tab:kronecker-correlation}
\begin{tabular}{@{}lr@{}}
\toprule
Statistic summed over the three partitions & Spearman correlation\\
\midrule
Number of distinct parts &0.6421\\
Durfee-square size &0.4857\\
\bottomrule
\end{tabular}
\end{table}

Table~\ref{tab:kronecker-correlation} compares two partition statistics on the same $n=10,11$ population, motivated by Pak--Panova's polynomial growth theorem, which assumes bounded Durfee-square sizes \citep{pakpanova2023}. The number of distinct parts correlates more strongly with the Kronecker coefficient than the Durfee-square size does. Pak--Panova's theorem bounds coefficient growth when Durfee-square sizes are bounded; the present observation compares individual coefficients empirically. However, it shows that, for the specific task of comparing two triples, the statistic in Table~\ref{tab:kronecker-comparison} is a better-performing choice than the one their proof uses.

\end{pipelineoutput}

\FloatBarrier
\Needspace{12\baselineskip}
\subsection{Character-table row sums}
\label{app:characters}
For a partition $\lambda\vdash n$, let $\chi^\lambda$ be the corresponding \emph{irreducible character} of $S_n$ and define
\[
R(\lambda)=\sum_{\rho\vdash n}\chi^\lambda(\rho).
\]
There is one summand per conjugacy class. Stanley's Problem~12 asks for a combinatorial interpretation of this integer; the representation-theoretic setting is described by \citet{sagan2001} and \citet{pmSundaram2018}. For the sign character of $S_3$, the three entries are $1,-1,1$, giving the sum $1$ in Figure~\ref{fig:characters}.

\begin{pipelineoutput}
\begin{figure}[H]
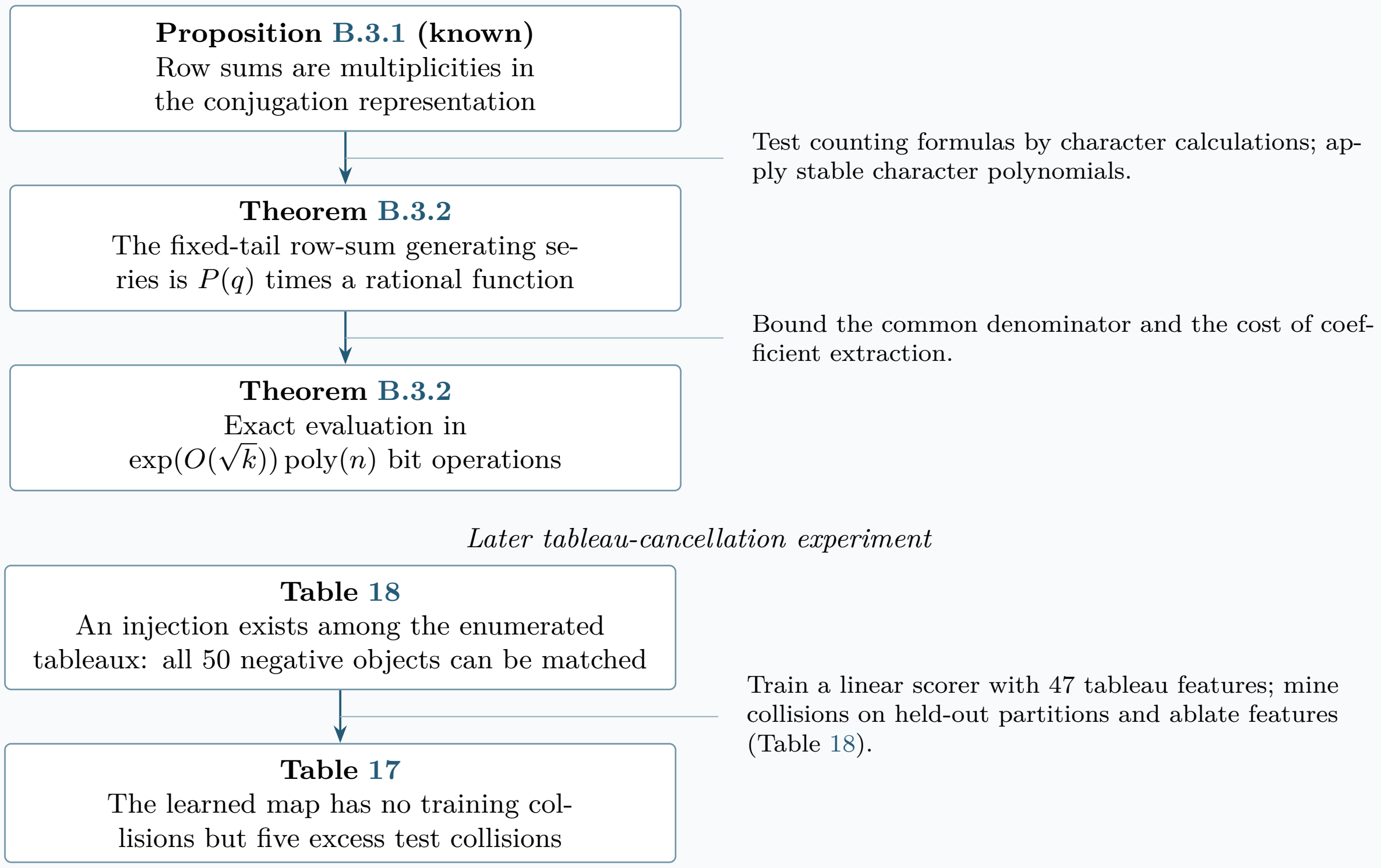

\centering
\begin{discoverychart}
\discoveryfirst{r1}{\textbf{Proposition~\ref{prop:row-conjugation} (known)}\\Row sums are multiplicities in the conjugation representation}
\discoverynext{r2}{r1}{\textbf{Theorem~\ref{thm:row-evaluation}}\\The fixed-tail row-sum generating series is $P(q)$ times a rational function}{Test counting formulas by character calculations; apply stable character polynomials.}
\discoverynext{r3}{r2}{\textbf{Theorem~\ref{thm:row-evaluation}}\\Exact evaluation in $\exp(O(\sqrt{k}))\operatorname{poly}(n)$ bit operations}{Bound the common denominator and the cost of coefficient extraction.}
\end{discoverychart}

\medskip
{\small\itshape Later tableau-cancellation experiment}\par\smallskip
\begin{discoverychart}
\discoveryfirst{r4}{\textbf{Table~\ref{tab:row-checks}}\\An injection exists among the enumerated tableaux: all 50 negative objects can be matched}
\discoverynext{r5}{r4}{\textbf{Table~\ref{tab:row-models}}\\The learned map has no training collisions but five excess test collisions}{Train a linear scorer with 47 tableau features; mine collisions on held-out partitions and ablate features (Table~\ref{tab:row-checks}).}
\end{discoverychart}
\caption{Character-row evaluation and the later search for a cancellation map.}
\label{fig:app-characters-flow}
\end{figure}

\begin{proposition}[{\citealp{pmSundaram2018}}]
\label{prop:row-conjugation}
The number $R(\lambda)$ is the multiplicity of the irreducible representation indexed by $\lambda$ in the conjugation action of $S_n$ on $\mathbb C[S_n]$. In particular, $R(\lambda)\geq0$.
\end{proposition}
This classical interpretation supplies the character decomposition underlying the tableau-cancellation experiments below.

\subsubsection{Models and cancellation experiments}
\label{app:row-learning}
The learning experiment sought an injection between finite sets of tableaux arising in the character decomposition. We used the expression
\begin{equation}
(s_{(2,2)}+s_{(2,1,1)})h_2[s_{(2,1)}]-s_{(1)}h_3[s_{(2,1)}],
\label{eq:row-cancellation}
\end{equation}
where brackets denote \emph{plethysm}. For $\nu\vdash10$ with $\nu_1\leq5$, we checked its $s_\nu$ coefficient against $[s_{(5,\nu)}]h_3[s_{(2,2,1)}]$. Such plethysms occur in the \emph{Frobenius characteristic} of the conjugation representation through its decomposition by cycle type \citep{pmSundaram2018}. Expanding the positive and negative terms of \eqref{eq:row-cancellation} gives the two sets to be matched; an injection would cancel the negative contribution.

A linear model scored each possible image using 47 integer-valued features of the tableaux, the added cell, and the summand contributing the source. Each source selected its highest-scoring image independently. We optimized the weights by 600 coordinate proposals from each of three initializations and selected the model on validation data. Output partitions were disjoint across the three splits, although the same constituent tableaux could occur in more than one split. The finite family had been explored symbolically before training; this split evaluates the learned scorer on held-out output partitions. We counted excess collisions as the number of sources minus the number of distinct images. The geometric baseline scores an image by $4a+2b-c+e$, where $a$ counts shared cells, $b$ counts agreeing entries, $c$ counts disagreeing entries, and $e$ indicates inclusion of the added cell. The lexicographic baseline chooses the first valid image in a fixed order.

\begin{table}[H]
\centering\small
\caption{Independent image selection in the tableau cancellation experiment.}
\label{tab:row-models}
\begin{tabular}{lrrrrr}
\toprule
& Sources & Output & \multicolumn{3}{c}{Excess collisions}\\
\cmidrule(lr){4-6}
Split & & partitions & Learned & Geometric & Lexicographic\\
\midrule
Training &22&10&0&8&12\\
Validation &11&4&1&1&7\\
Test &17&7&5&6&10\\
\bottomrule
\end{tabular}
\end{table}

The remaining nine output partitions have no negative
sources and require no matching. The selected rule was injective on every training output and on three of the seven test outputs. Its five test collisions exceeded the median of four from random linear scores (Table~\ref{tab:row-checks}). The random scores are an unselected reference distribution. Removing the feature measuring entry disagreement reduced test collisions while increasing validation collisions. Thus this experiment supplied failures of the proposed map and no improvement over the random-score comparison.

\begin{table}[H]
\centering\small
\caption{Checks and feature ablations for the cancellation experiment.}
\label{tab:row-checks}
\begin{tabularx}{\linewidth}{@{}>{\raggedright\arraybackslash}p{0.39\linewidth}Y@{}}
\toprule
Calculation & Result\\
\midrule
Tableau enumeration &97 positive and 50 negative objects over 30 output partitions.\\
Lookup injection &All 50 negative objects matched; 47 unmatched positive objects over 22 outputs.\\
Littlewood--Richardson enumeration &Two enumerations agree on all 240 ordered families.\\
Ambient plethysm coefficients &All 30 signed counts agree with Murnaghan--Nakayama and Jacobi--Trudi extraction.\\
100 random linear scores &Minimum, median, and maximum test collisions: $2,4,8$.\\
Entry-disagreement feature removed &Validation collisions $1\to4$; test collisions $5\to4$.\\
Added-cell entry feature removed &Test collisions $5\to6$.\\
\bottomrule
\end{tabularx}
\end{table}

The two Littlewood--Richardson enumerations used separate tableau predicates; the ambient coefficient calculations shared power-sum arithmetic. A concrete collision occurs at output $(4,3,3)$: sources obtained by adding cells $(3,3)$ and $(1,4)$ to shapes $(4,3,2)$ and $(3,3,3)$ select the same image. Both sources arise by adding a horizontal strip of three cells to $(3,3)$ before adding the final cell, and select the same positive tableau with inner shape $(2,2,2)$, content $(2,1,1)$, and reading word $1123$. The exhaustive matching in Table~\ref{tab:row-checks} establishes that an injection exists on this finite domain, while the learned rule fails to construct one.
\paragraph{Follow-up development experiments}
\label{app:row-development}
We subsequently reused all 50 negative objects as development
data, training a 47-parameter linear scorer and a
784-parameter tanh MLP with three seeds each.
One linear-model seed and two MLP seeds achieved
collision-free assignments. An integer rule with 22 nonzero coefficients,
extracted from the successful linear scorer, preserved
all 50 selected images.
These results concern the same 50 objects used for development.
A subsequent task audit identified a coordinated rank matching for unrestricted pairing within each output partition. The tested restriction allowing a one-cell change of the inner diagram admitted a greedy matching through nested neighbor sets.
Thus the development fits do not establish a learning
advantage over elementary combinatorial baselines.

\subsubsection{Evaluation for partitions with a fixed tail}
\label{app:row-evaluation}
The stable character polynomial gives an evaluation formula with complexity controlled by the number of boxes below the first row. Fix $\nu\vdash k$, $k\geq1$, and put $P(q)=\prod_{i\geq1}(1-q^i)^{-1}$ and $D_k(q)=\prod_{i=1}^k(1-q^i)$. Let $a_{\nu,n}$ be the sum over conjugacy classes of the virtual character with Frobenius characteristic $s_{(n-k,\nu)}$, defined by Jacobi--Trudi with $h_0=1$ and $h_j=0$ for $j<0$. For $n\geq k+\nu_1$, this is the ordinary row sum $R((n-k,\nu))$.

\begin{theorem}[LLM-generated]
\label{thm:row-evaluation}
There is an effectively computable polynomial $N_\nu(q)\in\mathbb Z[q]$ of degree at most $k(k+1)/2$ such that
\[
\sum_{n\geq0}a_{\nu,n}q^n=P(q)\frac{N_\nu(q)}{D_k(q)}.
\]
The numerator can be computed in $\exp(O(\sqrt{k}))\operatorname{poly}(k)$ bit time and space. The coefficients through degree $n$ can be computed in $\exp(O(\sqrt{k}))\operatorname{poly}(n)$ bit time. In particular, evaluation takes polynomial time in $n$ when $k=O((\log n)^2)$.
\end{theorem}
\begin{proof}
For $\gamma=(1^{m_1}2^{m_2}\cdots)\vdash k$, put $z_\gamma=\prod_i i^{m_i}m_i!$. The stable character polynomial of \citet[Proposition~I.1]{garsia2009} is
\[
\sum_{\gamma\vdash k}\frac{\chi^\nu(\gamma)}{z_\gamma}
\prod_i\sum_{a=0}^{m_i}(-1)^{m_i-a}i^a\binom{m_i}{a}(X_i)_a,
\]
where $X_i$ is the number of $i$-cycles and $(X)_a=X(X-1)\cdots(X-a+1)$. This polynomial also gives the generalized-Schur character below the ordinary range. Indeed, expanding the first row of Jacobi--Trudi gives an integer combination of permutation characters induced from Young subgroups valid for every $n$. It agrees with the stable polynomial for all sufficiently large $X_1$ and every nonnegative choice of the other cycle counts. Polynomial identity then gives equality in every degree.

Summing over cycle types amounts to substituting
\[
(X_i)_a\longmapsto\frac{a!q^{ia}}{(1-q^i)^a}
\]
after extracting the factor $P(q)$. Define
\[
B_{i,m}(q)=\sum_{a=0}^m(-1)^{m-a}i^a\binom ma a!q^{ia}(1-q^i)^{m-a},
\qquad G_\gamma(q)=\prod_i(1-q^i)^{m_i}.
\]
Then
\[
N_\nu(q)=\sum_{\gamma\vdash k}\frac{\chi^\nu(\gamma)}{z_\gamma}
\frac{D_k(q)}{G_\gamma(q)}\prod_i B_{i,m_i}(q).
\]
The multiplicity of the $h$th cyclotomic polynomial in $G_\gamma$ is $\sum_{h\mid i}m_i\leq\lfloor k/h\rfloor$, so $G_\gamma$ divides $D_k$ in $\Z[q]$. Since $\deg B_{i,m}\leq im$, each summand has degree at most $k(k+1)/2$. Integrality follows from the Jacobi--Trudi expansion in binomial polynomials: $\binom{X_i}{a}$ maps to $q^{ia}/(1-q^i)^a$, and every denominator divides $D_k$ by the same argument. Computation can therefore use integer arithmetic after multiplying by $k!$, followed by exact division.

Memoized Murnaghan--Nakayama computes the required character values using at most $(\sum_{j\leq k}p(j))^2$ pairs of partitions and polynomially many operations per pair. The bound $p(k)=\exp(O(\sqrt{k}))$ gives the stated number of operations. Character values, factorial scales, and polynomial coefficients have polynomial bit length in $k$: for example, $\lVert B_{i,m}\rVert_1\leq m!(i+2)^m$, and all polynomial degrees are $O(k^2)$. Writing $\delta=k(k+1)/2$, the coefficients of
$G_\gamma(q)^{-1}$ through degree $\delta$ are bounded
by $\binom{\delta+k}{k}$.
Since $\|D_k\|_1\leq 2^k$, the coefficients of
$D_k/G_\gamma$ also have polynomial bit length in $k$. This proves the numerator bound. Multiplying by $P(q)/D_k(q)$ through degree $n$ requires only $O(n^2)$ additions of integers with polynomial bit length, proving the coefficient bound.
\end{proof}
\end{pipelineoutput}

\FloatBarrier
\Needspace{12\baselineskip}
\subsection{Kazhdan--Lusztig coefficients}
\label{app:kazhdan-lusztig}
For $x\leq w$ in Bruhat order on $S_n$, let $P_{x,w}(q)$ be the Kazhdan--Lusztig polynomial, normalized by $P_{w,w}=1$ and $\deg P_{x,w}\leq(\ell(w)-\ell(x)-1)/2$ for $x<w$. Stanley's Problem~16 asks for a combinatorial interpretation of its coefficients. We use the conventions of \citet[Chapter~5]{bjornerbrenti2005}. Figure~\ref{fig:kazhdan-lusztig} shows $P_{1234,3412}(q)=1+q$.

Put $d=\ell(w)-\ell(x)$ and normalize the \emph{$R$-polynomial} by
\[
R_{x,w}(q)=q^{d/2}\widetilde R_{x,w}(q^{1/2}-q^{-1/2}).
\]
The polynomial $\widetilde R_{x,w}$ is monic of degree $d$, has nonnegative coefficients, and contains only powers of the same parity as $d$. For odd $d>0$, write $\mu(x,w)=[q^{(d-1)/2}]P_{x,w}$; set $\mu(x,w)=0$ otherwise. Combinatorial invariance asserts that $P_{x,w}$ depends only on the isomorphism type of the interval $[x,w]$ as a partially ordered set.

\begin{pipelineoutput}
\begin{figure}[H]
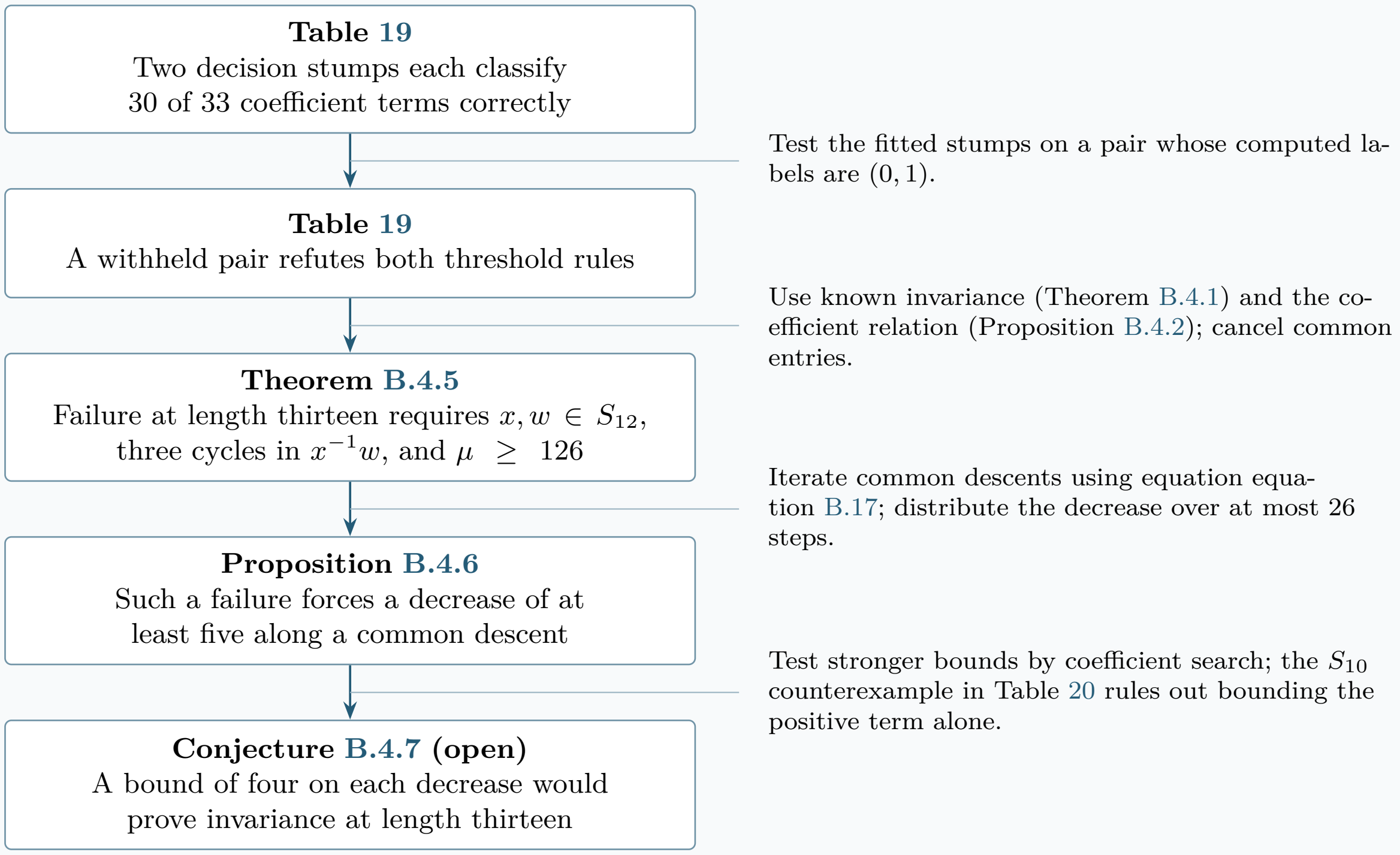

\centering
\begin{discoverychart}
\discoveryfirst{l1}{\textbf{Table~\ref{tab:kl-models}}\\Two decision stumps each classify 30 of 33 coefficient terms correctly}
\discoverynext{l2}{l1}{\textbf{Table~\ref{tab:kl-models}}\\A withheld pair refutes both threshold rules}{Test the fitted stumps on a pair whose computed labels are $(0,1)$.}
\discoverynext{l3}{l2}{\textbf{Theorem~\ref{thm:kl-twelve}}\\Failure at length thirteen requires $x,w\in S_{12}$, three cycles in $x^{-1}w$, and $\mu\geq126$}{Use known invariance (Theorem~\ref{thm:kl-known}) and the coefficient relation (Proposition~\ref{prop:kl-difference}); cancel common entries.}
\discoverynext{l4}{l3}{\textbf{Proposition~\ref{prop:kl-descent-reduction}}\\Such a failure forces a decrease of at least five along a common descent}{Iterate common descents using equation~\eqref{eq:kl-descent}; distribute the decrease over at most 26 steps.}
\discoverynext{l5}{l4}{\textbf{Conjecture~\ref{conj:kl-descent} (open)}\\A bound of four on each decrease would prove invariance at length thirteen}{Test stronger bounds by coefficient search; the $S_{10}$ counterexample in Table~\ref{tab:kl-counterexample} rules out bounding the positive term alone.}
\end{discoverychart}
\caption{Kazhdan--Lusztig counterexamples, proved reductions, and the remaining conjecture.}
\label{fig:app-kl-flow}
\end{figure}

\subsubsection{Models and counterexample experiments}
\label{app:kl-learning}
The learning experiment classified summands in a coefficient calculation as positive or zero. It used 33 terms from five permutation pairs after cancellation of common entries. The integer Kazhdan--Lusztig recurrence supplied the labels; Bruhat-order and descent tests selected eligible examples. Decision stumps, each using one feature threshold, were fitted to these terms and tested on a further pair with different labels (Table~\ref{tab:kl-models}). The two best-fitting rules both failed to distinguish that pair.

\begin{table}[H]
\centering\small
\caption{Decision stumps for vanishing of individual terms.}
\label{tab:kl-models}
\begin{tabularx}{\linewidth}{@{}>{\raggedright\arraybackslash}p{0.40\linewidth}Y@{}}
\toprule
Experiment & Result\\
\midrule
Training labels &16 positive and 17 zero terms from five permutation pairs.\\
Best decision stumps &Two tied rules, each with 3 errors among 33 terms.\\
Withheld pair &Labels $(0,1)$; both rules predict $(1,1)$.\\
\bottomrule
\end{tabularx}
\end{table}

\subsubsection{Known invariance and the possible coefficient difference}
\label{app:pm-kl}
\begin{theorem}[{\citealp[Corollary~1.3]{pmEMS2026}}]
\label{thm:kl-known}
For isomorphic type-A Bruhat intervals $I,I'$, one has $\widetilde R_I\equiv\widetilde R_{I'}\pmod{z^8}$. If their length is at most twelve, then $P_I=P_{I'}$.
\end{theorem}
This known invariance theorem determines the possible coefficient difference at length thirteen in Proposition~\ref{prop:kl-difference}.

\begin{proposition}[LLM-generated]
\label{prop:kl-difference}
For isomorphic type-A intervals $I,I'$ of length thirteen, there is an integer $\Delta$ such that
\begin{align}
\widetilde R_I(z)-\widetilde R_{I'}(z)&=\Delta z^9,\label{eq:kl-r-difference}\\
P_I(q)-P_{I'}(q)&=\Delta(q^2-9q^3+36q^4-84q^5+126q^6).
\label{eq:kl-discrepancy}
\end{align}
\end{proposition}
\begin{proof}
The coefficients below degree eight agree by Theorem~\ref{thm:kl-known}. The coefficient of $q$ in $P$ is invariant by \citet[Corollary~6.2]{pmPatimo}. Inversion gives $[z^{d-2}]\widetilde R_{x,w}=d+[q]P_{x,w}-\#\{y\in[x,w]:\ell(y)=\ell(w)-1\}$, so this coefficient is invariant as well. Monicity and parity leave only degree nine. Every proper subinterval has length at most twelve, so subtracting the two Kazhdan--Lusztig inversion identities gives
\[
q^{13}D(q^{-1})-D(q)=\Delta q^2(q-1)^9,
\qquad D(q)=P_I(q)-P_{I'}(q).
\]
Since $\deg D\leq6$, comparing coefficients through degree six proves \eqref{eq:kl-discrepancy}.
\end{proof}

\subsubsection{Reduction to twelve positions}
\label{app:pm-kl-frontier}
Cancellation removes entries that do not affect the interval. For $1\leq i,j\leq n$, define
\[
D_{x,w}(i,j)=\#\{k<i:w(k)\geq j\}-\#\{k<i:x(k)\geq j\}.
\]
If $x(i)=w(i)=j$ and $D_{x,w}(i,j)=0$, delete position $i$ and value $j$ from both permutations and relabel in increasing order. This preserves the interval length and $P_{x,w}$ \citep[Fact~8]{pmWarrington}. The induced interval identification preserves the same data on every subinterval, and hence also $R_{x,w}$ by inversion. Following Warrington, call a pair \emph{uncompressible} if no such deletion is possible.

\begin{lemma}[LLM-generated]
\label{lem:kl-factorization}
Label a saturated chain from $x$ to $w$ by its position transpositions and form a graph on $[n]$ with these transpositions as edges. Its connected components are independent of the chain. They give a product decomposition of the interval and of its $P$- and $\widetilde R$-polynomials. If $(x,w)$ is uncompressible, $d\geq3$ is odd, and $\mu(x,w)>0$, the graph is connected.
\end{lemma}
\begin{proof}
Saturated chains in a Bruhat interval are connected by changes across rank-two diamonds \citep[Chapter~2]{bjornerbrenti2005}. In type~A, the two transposition factorizations in a diamond involve either three positions or two disjoint pairs; the connected components are unchanged. Thus every vertex of the interval belongs to the coset of the reflection subgroup generated by transpositions within those components. Restricting the permutations to each component and relabeling gives the product of its symmetric groups. The induced Bruhat graph on the coset is the Bruhat graph of this reflection subgroup \citep[Theorem~2.2]{pmEMS2026}. Its interval is therefore a product. Increasing paths in the factors have a unique increasing interleaving for the restricted reflection order, so Dyer's path formula gives the product for $\widetilde R$ \citep[Theorem~5.3.4]{bjornerbrenti2005}. Inversion gives the product for $P$.

Two factors of positive length would force $\deg P_{x,w}\leq(d-2)/2$, contradicting $\mu(x,w)>0$. An isolated position $i$ is fixed throughout every saturated chain. A cover not involving $i$ cannot move a value across both position $i$ and value $x(i)$, since that would violate the type-A cover criterion. Hence the prefix count at $(i,x(i))$ is unchanged along the chain, so $D_{x,w}(i,x(i))=0$. This permits cancellation and contradicts uncompressibility.
\end{proof}

The number of cycles in the relative permutation gives a second constraint. For a word of $L$ transpositions on $m$ positions, let $k$ be the number of connected components of its graph and $c$ the number of cycles of its product, including fixed points. Then
\begin{equation}
G=\frac{L-m-c+2k}{2}\in\mathbb Z_{\geq0}.
\label{eq:kl-transpositions}
\end{equation}
Indeed, a transposition joining graph components decreases both $k$ and $c$ by one, a transposition splitting a cycle increases $c$ by one, and a transposition joining two cycles within a component decreases $c$ by one. These operations change $G$ by $0,0,1$, respectively. Reversing the word preserves $G$, so deleting either end cannot increase it. Extending a saturated chain proves $G(a,b)\leq G(x,w)$ for every subinterval; Lemma~\ref{lem:kl-factorization} makes these quantities independent of the chains.

\begin{lemma}[LLM-generated]
\label{lem:kl-position-bound}
Suppose an uncompressible pair $(x,w)$ in $S_m$ has odd length difference $d\geq3$ and $\mu(x,w)>0$. If $[z^{d-2r}]\widetilde R_{x,w}>0$, then $m\leq d-r+1$.
\end{lemma}
\begin{proof}
The graph is connected by Lemma~\ref{lem:kl-factorization}, so \eqref{eq:kl-transpositions} gives $c\leq d-m+2$. Dyer's formula supplies a path of $d-2r$ transpositions with product $x^{-1}w$. Any such product requires at least $m-c$ transpositions. Thus $m-c\leq d-2r$, and adding the two inequalities gives the bound.
\end{proof}

\begin{theorem}[LLM-generated]
\label{thm:kl-twelve}
If combinatorial invariance fails for type-A intervals of length thirteen, there is an uncompressible pair $x\leq w$ in $S_{12}$ with connected transposition graph such that
\[
\ell(w)-\ell(x)=13,\qquad
\#\operatorname{cycles}(x^{-1}w)=3,\qquad
[q^6]P_{x,w}\geq126.
\]
For this pair, $G(x,w)=0$.
\end{theorem}
\begin{proof}
Orient a pair of counterexamples so that $\Delta\geq1$ in Proposition~\ref{prop:kl-difference}. Nonnegativity gives $\mu\geq126\Delta$ and $[z^9]\widetilde R\geq\Delta$ for the first interval. After cancellation, Lemma~\ref{lem:kl-position-bound}, with $d=13$ and $r=2$, gives $m\leq12$. The largest leading coefficient through $S_{11}$ is 28 \citep[Theorem~1]{pmWarrington}, so $m=12$. The two inequalities in that lemma now give $c\leq3$ and $c\geq3$. Substitution in \eqref{eq:kl-transpositions} gives $G=0$.
\end{proof}

\subsubsection{A sufficient inequality along common descents}
For a common right descent $s$ of $x,w$ and odd $d\geq3$, the Kazhdan--Lusztig recurrence gives
\begin{equation}
\mu(x,w)-\mu(xs,ws)=[q^{(d-3)/2}]P_{x,ws}
-\sum_{\substack{x<z\leq ws,\ zs<z\\\ell(z)-\ell(x)\text{ odd}}}\mu(x,z)\mu(z,ws).
\label{eq:kl-descent}
\end{equation}
Here $P_{u,v}=0$ when $u\not\leq v$. To obtain this identity, take the coefficient of $q^{(d-1)/2}$ in the usual recurrence. The term $P_{xs,ws}$ contributes $\mu(xs,ws)$, the term $qP_{x,ws}$ contributes the displayed coefficient, and the degree bounds retain exactly the displayed correction terms.

\begin{proposition}[LLM-generated]
\label{prop:kl-descent-reduction}
If combinatorial invariance fails at length thirteen, there is an uncompressible pair in $S_{12}$ with length difference thirteen, connected transposition graph, three cycles in $x^{-1}w$, and $\mu(x,w)>0$, together with a common right descent $s$, for which
\[
\mu(x,w)-\mu(xs,ws)\geq5.
\]
\end{proposition}
\begin{proof}
Simultaneously applying a common descent preserves uncompressibility. For $s=(a,a+1)$, the prefix counts at common entries outside these two positions are unchanged. A common entry $j$ at position $a$ moves to $a+1$; both entries moved before it are less than $j$, so its cancellation condition is unchanged. A common entry at $a+1$ moves to $a$; both removed preceding entries exceed $j$, again leaving the difference unchanged. The relative permutation is conjugated by $s$, so its cycle count is preserved.

Start with the pair in Theorem~\ref{thm:kl-twelve}. Longest-element symmetry sends it to $(w_0w,w_0x)$ and preserves $\mu$ \citep[Fact~6]{pmWarrington}, uncompressibility, and the relative cycle count. The two lower lengths sum to $\binom{12}{2}-13=53$, so choose the orientation with $\ell(x)\leq26$. Whenever $\mu(x,w)>0$ and $d\geq3$, every right descent of $w$ is a descent of $x$ \citep[Fact~5]{pmWarrington}. Apply any such common descent. Uncompressibility and the cycle count persist, and Lemma~\ref{lem:kl-factorization} gives connectedness at each positive step.

Each step lowers $\ell(x)$ by one. The leading coefficient must vanish by the time $x$ is the identity, since the identity has no descent. Thus a sum of at most 26 successive differences in \eqref{eq:kl-descent} is at least 126. One difference is at least $\lceil126/26\rceil=5$.
\end{proof}

\begin{conjecture}[LLM-generated]
\label{conj:kl-descent}
For every uncompressible pair $x\leq w$ in $S_{12}$ with length difference thirteen, connected transposition graph, three cycles in $x^{-1}w$, and $\mu(x,w)>0$, every common right descent $s$ satisfies $\mu(x,w)-\mu(xs,ws)\leq4$.
\end{conjecture}
Proposition~\ref{prop:kl-descent-reduction} shows that this conjecture would imply combinatorial invariance at length thirteen.

A coefficient search ruled out a simpler bound on the first term of \eqref{eq:kl-descent}. The interval in Table~\ref{tab:kl-counterexample} has connected transposition graph and $G=0$, but its highest possible coefficient is eight. Therefore those two conditions alone do not bound this coefficient by four.

\begin{table}[H]
\centering\small
\caption{A counterexample to a coefficient bound based on connectedness and \eqref{eq:kl-transpositions}.}
\label{tab:kl-counterexample}
\begin{tabularx}{\linewidth}{@{}lY@{}}
\toprule
Quantity & Value\\
\midrule
Lower permutation $x$ &$(1,5,4,3,2,9,8,7,6,10)$\\
Upper permutation $v$ &$(5,7,8,1,9,10,2,3,4,6)$\\
Length difference &10\\
Cycles of $x^{-1}v$ &2\\
Transposition graph &Connected; $G(x,v)=0$.\\
Polynomial &$P_{x,v}=1+10q+29q^2+29q^3+8q^4$.\\
\bottomrule
\end{tabularx}
\end{table}
\end{pipelineoutput}

\FloatBarrier
\Needspace{12\baselineskip}
\subsection{Monomial immanants}
\label{app:immanants}
Let $h_r$ be the \emph{complete homogeneous symmetric function}, with $h_0=1$ and $h_r=0$ for $r<0$. For partitions $\nu\subseteq\mu$ padded to length $n$, the Jacobi--Trudi matrix of $D=\mu/\nu$ is
\[
H_D=(h_{\mu_i-\nu_j-i+j})_{1\leq i,j\leq n}.
\]
Let $\phi^\lambda$ be the class function of $S_n$ with Frobenius characteristic $m_\lambda$. The monomial immanant is
\[
\operatorname{Imm}_{\phi^\lambda}(H_D)=\sum_{w\in S_n}\phi^\lambda(w)\prod_{i=1}^n h_{\mu_i-\nu_{w(i)}-i+w(i)}.
\]
Stanley's Problem~18 asks whether its Schur coefficients are nonnegative. We follow \citet{pmLesnevich2024} for immanant characters and \citet{sagan2001} for symmetric functions. In Figure~\ref{fig:immanants}, $D=(2,1)$ and $\lambda=(2)$; the class function has values $0,2$, so the immanant is $2h_3=2s_{(3)}$.

\subsubsection{Character formulation}
\label{app:pm-immanants}
The Schur coefficients can be expressed through Kostka numbers. Put $N=|D|$, fix $\theta\vdash N$, and define
\[
B_i=\mu_i+n-i,\qquad C_j=\nu_j+n-j,\qquad
\alpha_j(w)=B_{w(j)}-C_j.
\]
Write $K_{\theta,\alpha}$ for the number of \emph{semistandard tableaux} of shape $\theta$ and content $\alpha$, and set it to zero if any entry of $\alpha$ is negative. Permuting the content leaves the Kostka number unchanged. For a conjugacy class $C_\rho\subseteq S_n$, define
\[
\Gamma_D^\theta(w)=\frac{n!}{|C_\rho|}\sum_{v\in C_\rho}K_{\theta,\alpha(v)}
\qquad(w\in C_\rho).
\]
This is Lesnevich's immanant character, with normalization chosen so that
\begin{equation}
[s_\theta]\operatorname{Imm}_{\phi^\lambda}(H_D)=\langle\Gamma_D^\theta,\phi^\lambda\rangle
\label{eq:immanant-pairing}
\end{equation}
for the usual character inner product \citep[Definition~2.8 and Lemma~2.9]{pmLesnevich2024}.

For a weakly increasing function $\eta:[n]\to[n]$ with $\eta(j)\geq j$, let $\widehat \eta(w)=\mathbf1\{w(j)\leq \eta(j)\text{ for all }j\}$. Its conjugacy average defines the Stanley--Stembridge character
\[
\Gamma_\eta(w)=\frac{n!}{|C_\rho|}\sum_{v\in C_\rho}\widehat \eta(v)
\qquad(w\in C_\rho).
\]
Our decomposition is first proved for the functions $K_{\theta,\alpha(w)}$ and $\widehat \eta(w)$, before averaging over conjugacy classes.

\begin{pipelineoutput}
\begin{figure}[H]
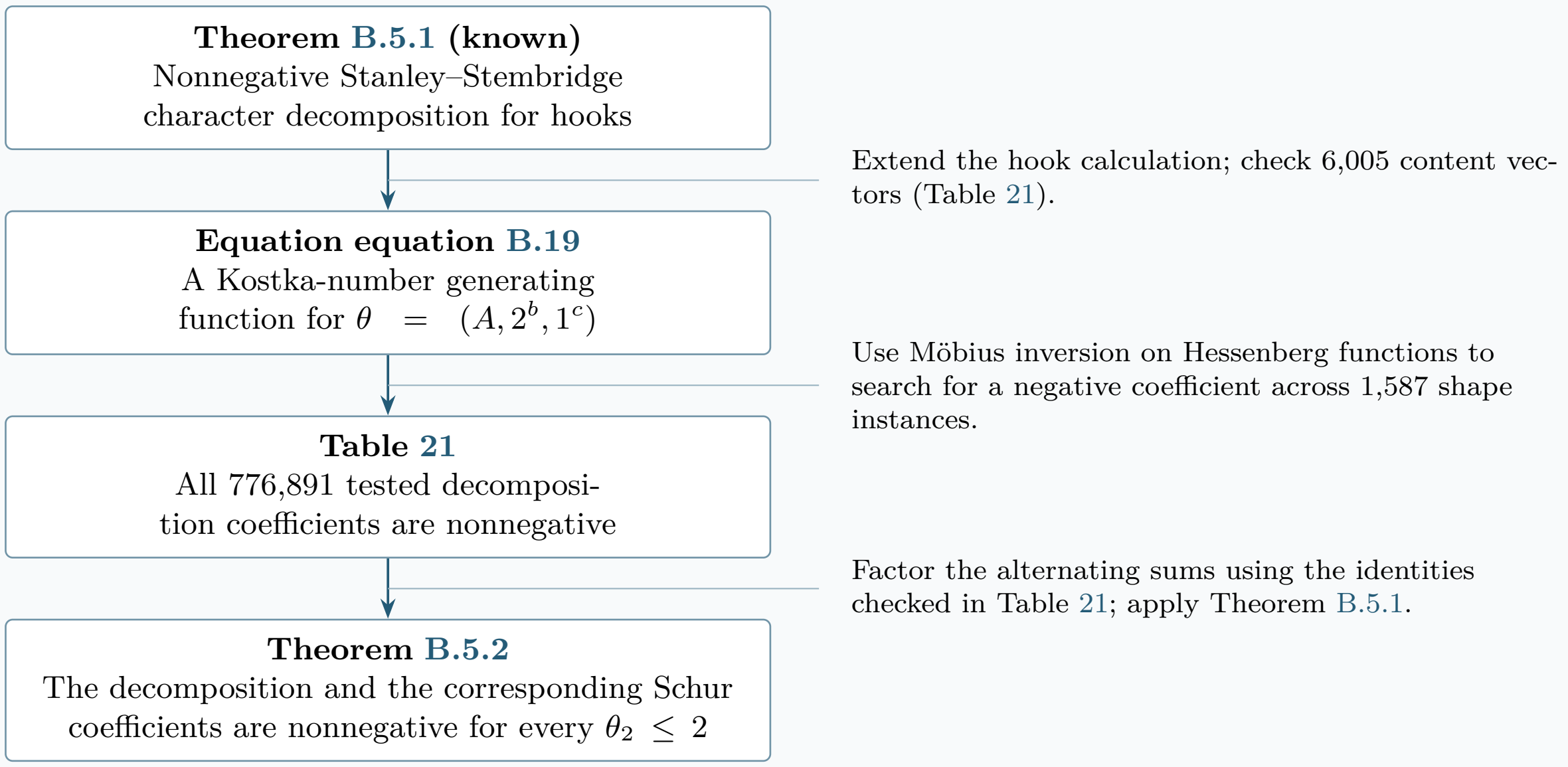

\centering
\begin{discoverychart}
\discoveryfirst{i1}{\textbf{Theorem~\ref{thm:immanant-known} (known)}\\Nonnegative Stanley--Stembridge character decomposition for hooks}
\discoverynext{i2}{i1}{\textbf{Equation~\eqref{eq:pm-content}}\\A Kostka-number generating function for $\theta=(A,2^b,1^c)$}{Extend the hook calculation; check 6,005 content vectors (Table~\ref{tab:immanant-experiments}).}
\discoverynext{i3}{i2}{\textbf{Table~\ref{tab:immanant-experiments}}\\All 776,891 tested decomposition coefficients are nonnegative}{Use M\"obius inversion on Hessenberg functions to search for a negative coefficient across 1,587 shape instances.}
\discoverynext{i4}{i3}{\textbf{Theorem~\ref{thm:immanant-positivity}}\\The decomposition and the corresponding Schur coefficients are nonnegative for every $\theta_2\leq2$}{Factor the alternating sums using the identities checked in Table~\ref{tab:immanant-experiments}; apply Theorem~\ref{thm:immanant-known}.}
\end{discoverychart}
\caption{Decomposition searches leading to immanant positivity.}
\label{fig:app-immanants-flow}
\end{figure}

\subsubsection{Decomposition experiments}
\label{app:immanant-checks}
The experiments enumerated Kostka numbers and tested expansions in the functions $\widehat \eta$. M\"obius inversion on the partially ordered set of Hessenberg functions determines the coefficients $a_\eta$ in
\[
K_{\theta,\alpha(w)}=\sum_\eta a_\eta\widehat \eta(w).
\]
A negative $a_\eta$ would refute this proposed decomposition. Table~\ref{tab:immanant-experiments} records the coefficient search and the checks of the identities used in the proof.

\begin{table}[H]
\centering\small
\caption{Checks for the immanant decomposition.}
\label{tab:immanant-experiments}
\begin{tabularx}{\linewidth}{@{}>{\raggedright\arraybackslash}p{0.49\linewidth}Y@{}}
\toprule
Test & Cases and outcome\\
\midrule
Content formula \eqref{eq:pm-content} &6,005 content vectors; all agree with Kostka numbers.\\
Path identities \eqref{eq:immanant-paths} &28 identities verified.\\
M\"obius coefficients $a_\eta$ &776,891 coefficients over 1,587 shape instances; all nonnegative.\\
Exponent in \eqref{eq:pm-pathproduct} &14,424 cases; all satisfy the stated bound.\\
Tableau-descent formula \eqref{eq:immanant-descents} &526 cases; all agree.\\
$D=(2^6)$, $\theta=(8,2,2)$ &38 positive coefficients $a_\eta$, with $\sum_\eta a_\eta=50$.\\
\bottomrule
\end{tabularx}
\end{table}

The final row counts distinct Hessenberg indices; their associated class functions need not be distinct. The identities checked here are used in the proof for every $\theta$ with $\theta_2\leq2$.

\subsubsection{Known positivity and its extension}
\begin{theorem}[{\citealp[Theorem~1.4]{pmLesnevich2024}; \citealp{pmHikita}}]
\label{thm:immanant-known}
If $\theta$ is a hook, then $\Gamma_D^\theta$ is a nonnegative integer combination of Stanley--Stembridge characters. Each Stanley--Stembridge character is a nonnegative integer combination of characters induced from the trivial representations of Young subgroups.
\end{theorem}
The two assertions are due to Lesnevich and Hikita, respectively. The induced characters have Frobenius characteristics $h_\lambda$, which are dual to the monomial functions $m_\lambda$. Thus \eqref{eq:immanant-pairing} is nonnegative whenever such a character decomposition exists.

\begin{theorem}[LLM-generated]
\label{app:pm-widthtwo}
\label{thm:immanant-positivity}
Let $D=\mu/\nu$ be a nonempty skew shape and $\theta\vdash|D|$ with $\theta_2\leq2$, taking a missing second part to be zero. Then $\Gamma_D^\theta$ is a nonnegative integer combination of Stanley--Stembridge characters. Consequently,
\[
[s_\theta]\operatorname{Imm}_{\phi^\lambda}(H_D)\geq0
\qquad\text{for every }\lambda\vdash n.
\]
If $D$ has no empty rows, there are integers $a_\eta\geq0$ satisfying the stronger identity $K_{\theta,\alpha(w)}=\sum_\eta a_\eta\widehat \eta(w)$ for every $w\in S_n$.
\end{theorem}
\begin{proof}
The column case follows from Theorem~\ref{thm:immanant-known}; write the remaining partitions as $\theta=(A,2^b,1^c)$ with $A\geq2$. We may assume $D$ has no empty rows. Removing empty rows and inducing back to the original symmetric group preserves a nonnegative Stanley--Stembridge decomposition \citep[Propositions~3.1 and~3.2]{pmLesnevich2024}.

\paragraph{A generating function for the Kostka numbers}
For a nonnegative content vector $\alpha$ of total size $A+2b+c$, the required Kostka number depends only on the number $r$ of positive entries and the number $o$ of entries equal to one. Set $U=(1+x)(1+y)$, $V=1+x+y$, $W=xy$. Then
\begin{equation}
 K_{(A,2^b,1^c),\alpha}=[x^{b+c}y^b](1-y/x)U^{r-o-1}V^o.
 \label{eq:pm-content}
\end{equation}
Indeed, for $\beta=(2^b,1^c)$, Bernstein creation gives $s_{(A,\beta)}=\sum_{j\ge0}(-1)^jh_{A+j}e_j^\perp s_\beta$. Dual Cauchy gives
\[
 \sum_{\gamma:\gamma_1\le2}(e_j^\perp s_\gamma)(X)s_{\gamma'}(x,y)=h_j(x,y)\prod_i(1+X_i x)(1+X_i y).
\]
After extracting $X^\alpha$, multiplication by the appropriate complete function contributes one for each monomial bounded coordinatewise by $\alpha$. The factors truncate to $1,V,U$ when $\alpha_i$ is zero, one, or at least two. Since $\sum_{j\ge0}(-1)^jh_j(x,y)=U^{-1}$, the coefficient of $s_{(b+c,b)}(x,y)$ is that in $U^{r-o-1}V^o$. In two variables this is the extraction in~\eqref{eq:pm-content}. A negative exponent means a formal power series with invertible constant term; only finitely many degrees contribute.

\paragraph{Prefix maxima and M\"obius inversion}
To express the Kostka number as a sum of Hessenberg indicator functions, group permutations by their prefix maxima. Write $p(w)_j=\max_{q\le j}w(q)$. At a nonrecord position $j$, an earlier record $\ell$ of value $m>w(j)$ gives
\[
 B_{w(j)}-C_j=(B_m-C_\ell)+(B_{w(j)}-B_m)+(C_\ell-C_j)\ge B_m-C_\ell+2.
\]
Thus nonnegativity is determined by the record positions and values; every nonrecord entry is at least two when the content is feasible. The prefix maxima therefore determine the numbers of zero and unit entries. Let $F(\eta)$ be the common Kostka number for permutations with $p(w)=\eta$, and set $F(\eta)=0$ for infeasible content. Since $\widehat \eta(w)=\mathbf1\{p(w)\le \eta\}$, the desired coefficients satisfy $F(k)=\sum_{\eta\ge k}a_\eta$. The distributive lattice of Hessenberg diagrams gives the local M\"obius formula
\begin{equation}
 a_\eta=\sum_{S\subseteq\mathcal A(\eta)}(-1)^{|S|}F(\eta+\mathbf1_S),\qquad \mathcal A(\eta)=\{i<n:\eta_i<\eta_{i+1}\}.
 \label{eq:pm-cube}
\end{equation}
Feasibility is $\eta_j\le H_j$, where $H_j=\max\{i:B_i\ge C_j\}$. If $\eta$ is infeasible its entire upper interval is infeasible. Otherwise retain only the corners in $\mathcal A_{\rm legal}(\eta)=\{i\in\mathcal A(\eta):\eta_i+1\leq H_i\}$; every subset of these increases is feasible.

\paragraph{Factorization over paths}
We factor the alternating sum in \eqref{eq:pm-cube} by recording how each allowed increase changes the content. Let $R,O$ be the initial numbers of positive and unit entries, and put $b_0=R-O$, the number of entries at least two. At a legal corner $i$, increasing $\eta_i$ changes the selected value at $i$ and, if the next record disappears, at that record. It either creates a zero, creates a one without creating a zero, or changes neither statistic. An increase that leaves both counts unchanged cancels the terms in~\eqref{eq:pm-cube} in pairs. Let $z$ count the corners at which the increase creates a unit entry.

Only the increases that create zeros can share an affected unit entry. Each removes at most two initial unit entries: the entry at $i$, if it was one, and the next record entry. The latter can be one only when it is at $i+1$ and $C_i-C_{i+1}=1$. Consequently these sets overlap only for adjacent corners, in one entry, and no entry belongs to three sets. Their overlap graph is a disjoint union of paths.

A component with $s$ corners has $s-1$ shared ones and $e\in\{0,1,2\}$ unshared endpoint ones. Denote by $c_0,c_2$ the numbers of components with $e=0$ and $e=2$, and by $s_{\rm tot}$ the total number of corners at which the increase creates a zero. The number of initial ones outside these sets is $f=O-s_{\rm tot}+c_0-c_2\ge0$.

For a component let $\operatorname{cov}(S)$ count its initial ones removed by a selected subset $S$ of corners, and put $h_j=h_j(x,y)$. The alternating contribution from this component is
\[
 M_{s,e}=\sum_{S\subseteq[s]}(-1)^{|S|}U^{s-|S|+\operatorname{cov}(S)}V^{s-1+e-\operatorname{cov}(S)}.
\]
These polynomials satisfy
\begin{align}
 M_{s,0}&=U^{s-1}(h_s+Wh_{s-1}),&M_{s,1}&=U^s h_s,\notag\\
 M_{s,2}&=U^s(h_s+h_{s+1}).&&
 \label{eq:immanant-paths}
\end{align}
To verify them for every $s$, replace $S$ by its complement $T$. Write $a,b\in\{0,1\}$ for the indicators for an unshared unit entry at each endpoint, so $e=a+b$. Then $M_{s,e}=(-1)^sU^{s-1+e}Z_s$, where $Z_s$ sums $(-U)^{|T|}(V/U)^{\#\text{uncovered ones}}$. Keeping the two states for membership of the last vertex in $T$ gives transfer matrix $\left(\begin{smallmatrix}1&1\\-U&-V\end{smallmatrix}\right)$, initial column $(1,-U(V/U)^a)^\mathsf{T}$, and terminal row $(1,(V/U)^b)$. Its characteristic polynomial is $(t+x)(t+y)$. Thus $(-1)^sZ_s$ obeys the complete-function recurrence; checking $s=1,2$ gives the identities at every length.

Each of the $z$ increases creating a unit entry contributes $1-V/U=W/U$. Combining components factors the generating function in the alternating sum as
\begin{equation}
 W^z U^{b_0-1-z-c_0}V^f\prod_{e=0}(h_s+Wh_{s-1})\prod_{e=1}h_s\prod_{e=2}(h_s+h_{s+1}),
 \label{eq:pm-pathproduct}
\end{equation}
where each factor uses its own component size. All factors are Schur-positive in two variables if $b_0-1-z-c_0\ge0$: $U=1+s_{(1)}+s_{(1,1)}$, $V=1+s_{(1)}$, $W=s_{(1,1)}$, and products preserve Schur positivity.

\paragraph{Nonnegativity of the remaining exponent}
It remains to prove that the exponent of $U$ in \eqref{eq:pm-pathproduct} is nonnegative when $b_0>0$. Each corner that creates a one, and the first corner of each component with $e=0$, uses a distinct initial entry at least two. The first corner of a component with $e=0$ cannot have its own initial one, since it would be unshared. Hence $b_0\ge z+c_0$.

One further entry is unused when $b_0>0$. Choose the rightmost position $j$ whose initial entry is at least two in a representative permutation. Every later position is a record, since every nonrecord has content at least two. Backward induction from the last record gives $w(\ell)=\eta_\ell=\ell$ for $\ell>j$. If $j$ is not a legal corner, its entry is unused in this assignment. Otherwise $\eta_j=j$, and the next initial entry $d=B_{j+1}-C_{j+1}$ is zero or one. Feasibility of the raise forces $d=1$, $C_j-C_{j+1}=1$, and raised entry zero. Thus $j$ removes the following one and does not create a one. If this unit entry is shared with the next corner, repeat the argument at that corner. The chain terminates at an unshared one because the last position has no corner. Its component therefore has $e\ge1$, so $j$ is not the assigned first corner of a component with $e=0$ either. This supplies one unused entry at least two and proves $b_0\ge z+c_0+1$. Equation~\eqref{eq:pm-pathproduct} is consequently Schur-positive whenever $b_0>0$.

\paragraph{The case of unit content}
When $b_0=0$, the coefficient has a direct description by standard tableaux. All content entries are then zero or one. Since there are $n$ nonempty rows, $N\ge n$, whereas $N=R=O\le n$. Hence $N=n$, all content entries are one, and all positions are records: $\eta=w=\mathrm{id}$ and $B_i=C_i+1$. A legal corner $i$ has $C_i-C_{i+1}=1$. Raising a subset $S$ merges consecutive chosen boundaries into blocks; the forward cyclic representative of each block has zero entries followed by its block length. After removing zeros the content is the composition with precisely those internal block boundaries $S$.

Write $\Des(T)=\{i:i+1\text{ lies in a lower row of }T\text{ than }i\}$. Standardization identifies the Kostka count for this composition with standard tableaux whose descent set avoids $S$. Inclusion--exclusion gives
\begin{equation}
 a_{\mathrm{id}}=\#\{T\in\operatorname{SYT}(\theta):\mathcal A_{\rm legal}(\mathrm{id})\subseteq\Des(T)\}\ge0.
\label{eq:immanant-descents}
\end{equation}
Together with~\eqref{eq:pm-pathproduct}, this proves $a_\eta\geq0$ for every Hessenberg function $\eta$. Averaging over conjugacy classes proves the character decomposition. The nonnegativity of the immanant coefficient follows from \eqref{eq:immanant-pairing} and Theorem~\ref{thm:immanant-known}.
\end{proof}
\end{pipelineoutput}

\subsection{Recovered results and generated conjectures}
\label{app:conjecture-history}

\subsubsection{Recovering known statements}

The descent and essential-rank conditions in Propositions~\ref{prop:schubert-descent} and~\ref{prop:schubert-rank} were formulated during the Schubert investigations and identified as consequences of established theory. Similarly, the Kronecker experiments recovered Murnaghan's inequalities and Rosas's bounds. The semigroup inequality was instead supplied as a known relation against which to test predictions. These are different uses of prior mathematics.

\paragraph{Hook row sums}
\label{app:recovered-row-formulas}
The row-sum investigation reconstructed a positive formula for hook partitions. For a cycle type $\rho$, let $Z_\rho$ be the centralizer of a permutation of that type and let $\mathrm{Std}_n$ be the standard representation. The conjugation interpretation gives
\[
 R((n-r,1^r))=\sum_{\rho\vdash n}
 \dim\bigl(\mathop{\bigwedge}\nolimits^r\mathrm{Std}_n\bigr)^{Z_\rho}.
\]
On each cycle, a Fourier basis identifies invariant exterior monomials by subsets whose indices sum to zero modulo the cycle length. Taking invariants under permutations of equal-length cycles and removing the global trivial summand gives the positive counting formula. The implementation agrees with all 55 hook row sums through $n=10$ and with direct centralizer calculations for all 240 hook multiplicities over the 44 cycle types through $n=7$. This reconstructs known hook formulas; the classical representation-theoretic background is described by \citet{pmSundaram2018}. These checks are separate from the fixed-tail evaluation theorem and the learned tableau map.

\paragraph{Invariance through length twelve}
\label{app:kl-rederivation}
The Kazhdan--Lusztig investigation also derived the bound now stated in Theorem~\ref{thm:kl-known}. The low-degree congruence modulo $z^8$, parity, monicity, and invariance of $[z^{d-2}]\widetilde R$ determine all coefficients through interval length eleven. The last invariance follows from the invariant linear coefficient of $P$ and the coatom count, as in the proof of Proposition~\ref{prop:kl-difference}. For an even interval length $d>0$, the standard inversion identity gives
\[
 2R_{x,w}=-\sum_{x<z<w}(-1)^{\ell(z)-\ell(x)}R_{x,z}R_{z,w}.
\]
Every factor on the right belongs to a proper subinterval. Hence invariance through length eleven implies invariance of $R$ at length twelve; Kazhdan--Lusztig inversion and the degree bound then give invariance of $P$. This rederives the established invariance result. The length-thirteen restrictions in Theorem~\ref{thm:kl-twelve} are separate deductions.

\subsubsection{Two remaining Kazhdan--Lusztig conjectures}

In addition to the corrected-descent conjecture, the investigation proposed a stronger coefficient bound. Use the uncompressibility and connected transposition graph defined in Appendix~\ref{app:pm-kl-frontier}. If the interval has $m$ positions, odd length $d$, and $c$ cycles in $x^{-1}w$, including fixed points, put
\[
 G(x,w)=\frac{d-m-c+2}{2}.
\]
The transposition argument in~\eqref{eq:kl-transpositions} gives $G\geq0$.
\begin{conjecture}
\label{conj:kl-genus-zero}
For every uncompressible type-A interval $[x,w]$ with connected transposition graph and odd length $d$, if $G(x,w)=0$, then
\[
 [q^{(d-1)/2}]P_{x,w}(q)\leq1.
\]
\end{conjecture}
This concerns the coefficient at the largest permitted degree, which may be zero even when the polynomial has a nonzero coefficient in a smaller degree. The hypothetical obstruction in Theorem~\ref{thm:kl-twelve} has $(m,d,c)=(12,13,3)$, so this conjecture would exclude it. Conjecture~\ref{conj:kl-descent} instead bounds a corrected difference on that specific length-thirteen class. Both remain open. The even-length coefficient-eight example in Table~\ref{tab:kl-counterexample} refutes the broader bound that omits recurrence corrections. Bounded examples and the standard small-rank zero--one results do not establish the new conjecture.

\subsubsection{From the first non-hook immanant case to width two}
\label{app:immanant-conjecture-history}

For a feasible content vector $\alpha$ of total size $N\geq4$, let $r$ count its positive entries and let $t$ count entries at least two. The first non-hook calculation gives
\[
 K_{(N-2,2),\alpha}=\binom{r-1}{2}+t-1.
\]
The binomial term is the known hook weight for $(N-2,1,1)$. Searches suggested that the complete weight admits a nonnegative pointwise expansion in the indicators $\widehat\eta$. A stronger attempted shortcut, requiring the correction $\delta=t-1$ to have such an expansion by itself, fails for the skew shape $(3,3,3,1)/(1,1)$. With $\delta=0$ on infeasible contents, its unique expansion is
\[
 \delta=\widehat{(1,3,4,4)}+\widehat{(2,2,4,4)}
   -\widehat{(2,3,4,4)}+\widehat{(3,3,4,4)}.
\]
Here $\widehat{(\eta_1,\ldots,\eta_4)}$ denotes the indicator associated with the displayed Hessenberg function, using one-based indices. The negative coefficient led to a revised cancellation conjecture: each negative correction coefficient is $-1$ and occurs at an index where the hook contribution is at least one. A local analysis of record positions proved this cancellation, establishing the $(N-2,2)$ case.

Subsequent extensions to $(A,2,1^c)$ suggested the general statement: for every skew shape and every coefficient partition $\theta$ with $\theta_2\leq2$, the full Kostka weight has a nonnegative pointwise expansion in Hessenberg indicators. This generated conjecture is now Theorem~\ref{thm:immanant-positivity}, proved by the content series, path factorization and boundary argument in Appendix~\ref{app:immanants}. It allows arbitrarily many parts equal to two. Lesnevich's hook theorem was the starting input, and positivity of the Stanley--Stembridge summands converts the decomposition into Schur positivity. The conjecture-to-proof trajectory used computation and symbolic reasoning; the separate classifier below addressed another question.

\subsection{Additional cross-problem analysis}
\label{app:cross-study-controls}

\paragraph{Schubert predicate extraction without a model}
A retrospective comparison repeated the historical predicate searches on the
recorded data using 37 curated features, without model predictions. It
recovered the descent condition and the two essential-rank predicates.
Among single-feature candidates, the both-tight essential-rank condition
ranked first by the number of excluded zeros. Validation on 83,689 separate
positive examples removed the spurious candidate regions. This comparison
conditions on features developed during the earlier investigation; it does
not measure discovery from an independently specified initial representation.
The random-feature comparison in Table~\ref{tab:schubert-features} likewise
uses activations of the trained model for both representations.

\paragraph{Later row-sum matching experiments}
Appendix~\ref{app:row-development} reports the fits on all 50 negative objects and the coordinated matching constructions found in the subsequent task audit.

\paragraph{Classifying enumerated skew-shape targets}
Each example is a vector of unscaled sums of Kostka weights over conjugacy
classes of $S_6$, with identity coordinate 24. The two inputs are its
coordinates for cycle types $(2,1^4)$ and $(5,1)$. Positive examples come from
the enumerated skew shapes; negative examples are constructed vectors absent
from that collection. All examples have positive five-cycle coordinate.
The labels concern membership in this collection. The stump fits 550/550 training examples and scores
279/279 on an ordinary holdout containing 268 vectors from enumerated shapes and 11 constructed negative vectors.
It scores 0/15 on a mined challenge containing only constructed negative vectors. After
inspection of that challenge, adding 12 training examples from different
parents and fitting a depth-two tree gives 562/562 on training, 278/279 on the
ordinary holdout, and 14/15 on the reused challenge. Both models fail the
recorded zero-error criterion. The changed training data and depth are not a
controlled architecture comparison, and this study is separate from the
proof of Theorem~\ref{thm:immanant-positivity}.

\FloatBarrier
\clearpage
\subsection{Small-model analyses for Problems 16 and 18}
\label{app:small-model-plots}

Both investigations combined LLM-guided mathematical arguments with small-model experiments. The plots below show how fitted rules, hard negatives, and branch tests informed the model analyses. The recorded data and plotting code are available in the \href{https://github.com/uw-math-ai/PermuFormer2/tree/main/Stanley-Problem-16/graphs}{Problem 16 figures} and \href{https://github.com/uw-math-ai/PermuFormer2/tree/main/Stanley-Problem-18/graphs}{Problem 18 figures} directories.

\begin{pipelineoutput}
\paragraph{Kazhdan--Lusztig recurrence terms}
The two fitted stumps in Table~\ref{tab:kl-models} proposed conditions for a recurrence term to be positive. One used the number of atoms of the associated Bruhat interval; the other tested whether the cover transposition exchanges adjacent positions or adjacent values. The withheld pair has the same values of these two features and different labels, refuting both conditions. Figure~\ref{fig:kl-model-assessment} compares their predictions with the training-majority baseline.

\begin{figure}[H]
\centering
\includegraphics[width=\linewidth]{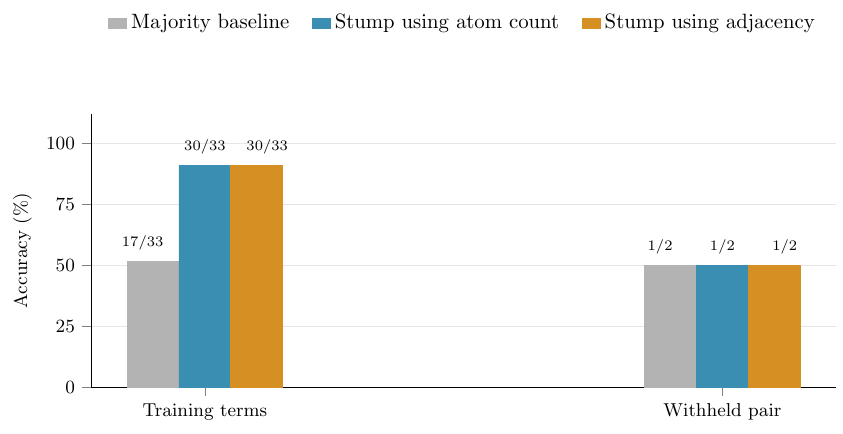}
\caption{\textbf{Fitted stumps and the withheld pair.} Both stumps classify 30 of 33 training terms correctly, compared with 17 for the majority baseline. Each predicts one of the two withheld labels correctly. The pair was selected to test the learned conditions, so its accuracy does not estimate population performance.}
\label{fig:kl-model-assessment}
\end{figure}

\paragraph{Character data from skew shapes}
The classifier in Appendix~\ref{app:cross-study-controls} tested membership in an enumerated collection of character-data vectors. Hard-negative mining found 15 constructed vectors that the initial stump classified as members. Adding 12 training examples and fitting a depth-two tree corrected 14 of these errors (Figure~\ref{fig:immanant-model-assessment}). The same 15 vectors were reused after inspection. These experiments studied the learned membership rule; Theorem~\ref{thm:immanant-positivity} follows from the decomposition calculations and symbolic argument in Appendix~\ref{app:immanants}.

\begin{figure}[H]
\centering
\includegraphics[width=\linewidth]{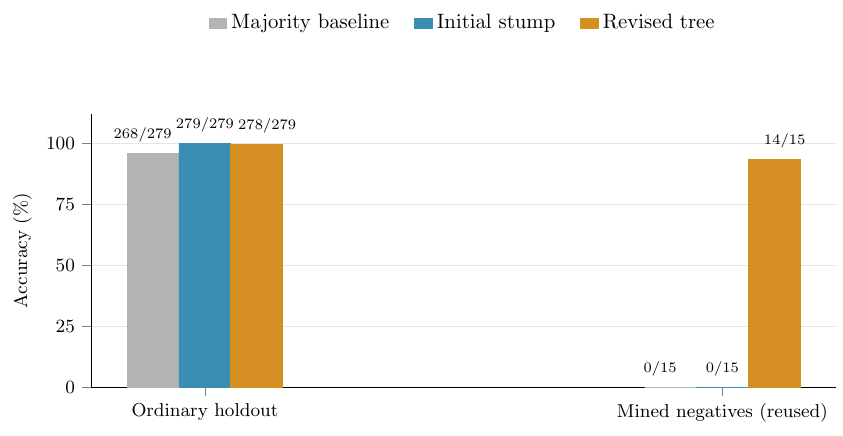}
\caption{\textbf{Hard negatives expose errors in shape-membership predictions.} The ordinary holdout contains 268 vectors from enumerated shapes and 11 constructed negatives. The majority baseline always predicts membership. Training data and tree depth both changed in the revision.}
\label{fig:immanant-model-assessment}
\end{figure}

\begin{figure}[H]
\centering
\includegraphics[width=\linewidth]{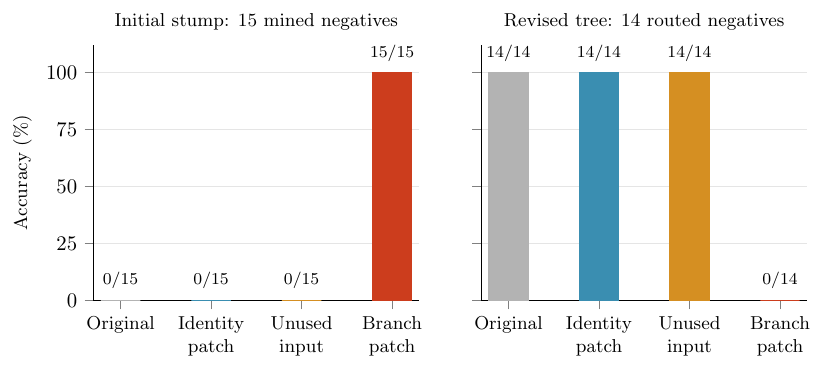}
\caption{\textbf{Branch tests identify which decisions control the predictions.} Replacing the stump's branch decision with that from a negative training example corrects all 15 mined cases. Replacing the revised tree's child decision with that from a positive training example makes all 14 cases reaching that node incorrect. Identity patches and changes to unused inputs preserve the original predictions. The two panels use different donor classes and sample counts.}
\label{fig:immanant-branch-tests}
\end{figure}
\end{pipelineoutput}
\section{Direct LLM baselines}
\label{app:direct-llm-baselines}

\paragraph{Protocol.}
We requested \texttt{gpt-6.2-astra} and \texttt{gpt-6.2-sol} through the Codex CLI. Each received a fresh context, a separate workspace, the statement of one Stanley problem, and the same procedural instructions. Agents could choose their own subquestions, write exact computations using Python and its standard library, search for counterexamples, and attempt proofs. The prompt \textbf{excluded} the manuscript results, PatternMiner workflow, earlier outputs, online retrieval, subagents, and predictor training. It allowed prior mathematical knowledge and asked agents to identify recalled results. The supervisor launched, timed, and logged the runs without mathematical follow-up. The 900-second allowance included computation and time waiting for model and tool responses.

\paragraph{Outcomes at 15 minutes.}
Table~\ref{tab:direct-llm-baselines} gives the running times and selected outcomes of both attempts on each problem. Completion records whether the process finished within its allowance. After freezing the outputs, separate agents reviewed all 66 submitted claim records, checked arguments, reran computations, and examined selected prior work. The reviews accepted the arguments for 53 records within their stated domains, classified nine as finite evidence, and left four unproved in the timed submissions.

\begin{table}[htbp]
\centering\small
\caption{Direct baseline attempts at 15 minutes. Times are elapsed seconds. The final column gives selected mathematical outcomes.}
\label{tab:direct-llm-baselines}
\begin{tabularx}{\linewidth}{@{}lrrY@{}}
\toprule
Problem & Astra & Sol & Outcome \\
\midrule
Kronecker (10) & 900$^*$ & 890 & Global matching rule for coefficients with a two-row index. Testing membership can require constructing large lists. \\
Schubert (11) & 876 & 880 & Positive rules for restricted output permutations or separated input supports. The two agents treated different families. \\
Row sums (12) & 860 & 859 & Unsigned matching count for all shapes and direct rules for hooks and two-row shapes. The all-shape construction uses an algebraic argument and has no established efficient membership test. \\
Kazhdan--Lusztig (16) & 891 & 875 & Coefficient formulas for restricted interval families, including Boolean-middle intervals with specified endpoints. Classical formulas cover some of the cases. \\
Immanants (18) & 900$^*$ & 898 & Both agents gave formulas for matrices of lower bandwidth at most one. Positivity on this family also follows from prior work. \\
\bottomrule
\end{tabularx}
\par\smallskip\raggedright $^*$Both Astra runs reached the 15-minute limit. We then made one fresh 30-minute attempt on each problem. The Kronecker run finished in 29 min 13 s and gave a block formula under a row-gap condition. The immanant run finished in 29 min 40 s and gave a positivity formula for matrices of lower bandwidth at most two with nonzero second-subdiagonal entries in nonconsecutive rows.
\end{table}

\paragraph{Matched Schubert comparison.}
In a separate comparison on the Schubert projection problem, two GPT-6.1 Sol extra-high agents received the same data, tools, 60-minute budget, and allowance of 50,000 coefficient or projection queries, with prompts differing only by the instruction to train and use a small model. Both found zeros satisfying $\Des(w)\subseteq\Des(u)\cup\Des(v)$ for which the smallest number of retained flag dimensions giving an empty generic intersection was four, five, and six in $S_7$, $S_8$, and $S_9$, respectively. Without the training instruction, the first certificates in the query logs used 438, 7,112, and 18,591 queries, compared with 78, 784, and 2,953 for the full workflow. This secondary measure was chosen after seeing interim progress and before computing the counts, with one agent session per condition.

\end{document}